\documentclass[twoside]{article}

\usepackage[accepted]{aistats2022}
\usepackage[utf8]{inputenc} 
\usepackage[T1]{fontenc}    
\usepackage{hyperref}       
\usepackage{url}            
\usepackage{booktabs}       
\usepackage{amsfonts}       
\usepackage{nicefrac}       
\usepackage{microtype}      
\usepackage{xcolor}         
\usepackage{xspace}

\usepackage[round]{natbib}

\usepackage{amsfonts}
\usepackage{amsmath}
\usepackage{amssymb}
\usepackage{bm}
\usepackage{graphicx}
\usepackage{float}
\usepackage{hyperref}
\usepackage{tabularx}
\usepackage{balance}
\usepackage[noend]{algpseudocode}
\usepackage{algorithmicx, algorithm}
\makeatletter
\renewcommand{\ALG@beginalgorithmic}{\small}
\algrenewcommand\alglinenumber[1]{\scriptsize #1:}
\makeatother

\usepackage{enumitem}
\setlist{itemsep=1pt,topsep=2pt,leftmargin=*}
\newcommand{\E}{\mathbb{E}}
\DeclareMathOperator*{\EE}{\mathbb{ E}}

\DeclareMathOperator*{\Var}{Var}
\DeclareMathOperator{\KL}{KL}
\DeclareMathOperator{\diag}{diag}
\newcommand{\pMPF}{\hat{p}_{\scriptscriptstyle \text{MPF}}}

\newcommand{\sss}[1]{\scriptscriptstyle{\text{#1}}}
\newcommand{\VSIS}{IWVI\xspace}
\usepackage{textcomp}

\usepackage{amsthm}
\usepackage{multirow}
\usepackage{tablefootnote}
\usepackage{threeparttable}
\newtheorem{definition}{Definition}
\newtheorem{theorem}{Theorem}
\newtheorem{lemma}{Lemma}

\makeatletter
\newcommand{\thickhline}{%
    \noalign {\ifnum 0=`}\fi \hrule height 1pt
    \futurelet \reserved@a \@xhline
}

\newcolumntype{"}{@{\hskip\tabcolsep\vrule width 1pt\hskip\tabcolsep}}
\makeatother

\begin{document}

%

%

\twocolumn[

\aistatstitle{Variational Marginal Particle Filters}

\aistatsauthor{ Jinlin Lai \And Justin Domke \And  Daniel Sheldon }

\aistatsaddress{Manning College of Information and Computer Sciences, University of Massachusetts Amherst \\
\texttt{\{jinlinlai,domke,sheldon\}@cs.umass.edu} }
]

\begin{abstract}
Variational inference for state space models (SSMs) is known to be hard in general. Recent works focus on deriving variational objectives for SSMs from unbiased sequential Monte Carlo estimators. We reveal that the marginal particle filter is obtained from sequential Monte Carlo by applying Rao-Blackwellization operations, which sacrifices the trajectory information for reduced variance and differentiability. We propose the variational marginal particle filter (VMPF), which is a differentiable and reparameterizable variational filtering objective for SSMs based on an unbiased estimator. We find that VMPF with biased gradients gives tighter bounds than previous objectives, and the unbiased reparameterization gradients are sometimes beneficial.
\end{abstract}

\section{Introduction}
Sequential data are often described by state space models (SSMs), where the latent variables $x$ and the observed variables $y$ vary over time. An SSM is defined as
\begin{equation}
  p(x_{1:T},y_{1:T})=f(x_1)\prod_{t=2}^Tf(x_{t}|x_{t-1}) \prod_{t=1}^Tg(y_t|x_t).
  \label{SSM}
\end{equation}
With this model, given one or more observed sequences $y_{1:T}$, two tasks are of interest:
\begin{itemize}
	\item What is the closest distribution $q(x_{1:T};\phi)$ to the posterior distribution $p(x_{1:T}|y_{1:T})$?
	\item What are the best parameters $\theta$ for $p(x_{1:T}, y_{1:T})$ to model the observed data?
\end{itemize}
These two tasks can be simultaneously solved by variational inference (VI) \citep{Blei2016VariationalIA}. 
Recent advances \citep{naesseth2018variational,maddison2017filtering,le2017auto} improve the performance of VI on SSMs by using sequential Monte Carlo (SMC) or the particle filter (PF) \citep{naesseth2019elements,doucet2009tutorial} to define a variational objective, which gives \emph{variational} sequential Monte Carlo (VSMC). 
However, the SMC resampling step is problematic: sampling of discrete ancestor variables is non-differentiable, and thus reparameterization gradient estimators \citep{kingma2013auto,rezende2014stochastic} cannot be used. 
Most practical implementations use a biased gradient estimator instead.
Recently, \citet{corenflos2021differentiable} proposed a fully differentiable particle filter, which gives up the guarantee of being a lower bound but approximates VSMC well.

This paper makes three contributions. First, we give a novel proof of the correctness of SMC in terms of transformations of estimator-coupling pairs \citep{domke2019divide}.
This provides a high-level view of SMC that complements existing perspectives.
Second, we marginalize (Rao-Blackwellize) over the choice of ancestor variables in each step to get the marginal particle filter (MPF) \citep{klaas2012toward}, which gives a novel proof of correctness of MPF and reveals the direct relationship between MPF and SMC.
Third, we propose to optimize a variational bound based on MPF, which we call the variational marginal particle filter (VMPF). Since Rao-Blackwellization reduces the variance of estimators, we expect that the VMPF bound is tighter than VSMC and leads to better inference and learning. Furthermore, we observe that MPF can be rephrased without discrete ancestor variables as sampling from mixtures.
As it is often possible to efficiently differentiate through mixtures (currently, this is possible for Gaussian or product distributions)
\citep{graves2016stochastic,figurnov2018implicit}, {\em unbiased} reparameterization gradients can be computed for the VMPF when suitable proposal distributions are used.

We conduct experiments using the VMPF on linear Gaussian SSMs, stochastic volatility models and deep Markov models \citep{KrishnanSS17}. Our first aim is to understand the significance of unbiased gradients. We confirm that these do indeed lead to tighter bounds given sufficient optimization, especially with higher numbers of particles, but that a biased gradient estimator typically performs better on complex models due to lower gradient variance. Our second aim is to understand the significance of the lower variance of the MPF vs SMC for estimating the log-likelihood. We find that, as expected, this manifests as the VMPF bound being tighter than the VSMC bound, which implies better posterior distributions and better models.


We note that, while marginalizing ancestor variables is often beneficial, it prevents applying VMPF to some models, such as VRNNs~\citep{ChungKDGCB15}, which require access to those variables (see Section~\ref{sec:limitations}). 
The time complexity of VMPF with $N$ particles is $O(N^2)$ (compared to $O(N)$ for VSMC), but in many practical situations the $O(N^2)$ component is dominated by an $O(N)$ component with higher constants.




\section{Background}
\textbf{Variational Inference and Couplings.} Variational inference 
seeks to produce an approximation $q(x_{1:T};\phi)\approx p(x_{1:T}|y_{1:T})$ given a joint distribution $p(x_{1:T}, y_{1:T})$ such as the SSM in Equation~\eqref{SSM} and fixed observations $y_{1:T}$. It is based on the fact that minimizing the KL divergence between the two distributions is equivalent to maximizing the evidence lower bound (ELBO) \citep{JordanGJS99} $\mathcal{L}_{\scriptscriptstyle \text{ELBO}}(\phi)=\mathbb{E}_{q(x_{1:T};\phi)}[\log p(x_{1:T},y_{1:T})-\log q(x_{1:T};\phi)]$.
The ELBO also lower bounds $\log p(y_{1:T})$, which is the basis of variational expectation maximization (VEM) \citep{bernardo2003variational} for learning model parameters $\theta$. This paper will use a generalization of the VI framework based on estimator-coupling pairs \citep{domke2019divide} , which define a family of
algorithms that have the above properties automatically. This is easier to describe with a more general notation: let $\gamma(x)$ be an unnormalized distribution with normalizer $Z = \int \gamma(x)dx$, so the goal is to approximate $\pi(x) = \gamma(x)/Z$. (For Bayesian inference, $\gamma(x)=p(x,y)$, $\pi(x)=p(x \vert y)$ and $Z=p(y)$, where the evidence $y$ is fixed.) 
\begin{definition}
\label{definition1}
An estimator $R(\omega) > 0$ and a distribution $a(x|\omega)$ are a valid estimator-coupling pair
for $\gamma(x)$ under distribution $Q(\omega)$ if, for all $x$,
	\begin{align*}
		\mathbb{E}_{Q(\omega)}[R(\omega)a(x|\omega)]=\gamma(x).
	\end{align*}
\end{definition}
This implies several things. First, $\E_{Q(\omega)} [R(\omega)] = Z$, meaning $R$ is an unbiased estimator of the normalizer $Z$. Second, more generally, $\E_{Q(\omega)a(x|\omega)}[R(\omega)f(x)] = \int f(x) \gamma(x) dx$ for any (integrable) test function $f$. Third, $\pi(\omega,x)=\frac{1}{Z}Q(\omega)R(\omega)a(x|\omega)$ is a ``coupling'': a distribution that has both $\frac{1}{Z}Q(\omega)R(\omega)$ and $\pi(x)$ marginals.
From the fact that $\E[R] = Z$, it is easy to see that $\mathbb{E}[\log R]\leq \log Z$. The looseness of this bound can be decomposed as
\begin{align}
	\label{equation3}
	\log Z =& \mathbb{E}_{Q(\omega)}[\log R(\omega)]+\notag\\
	 & \KL(Q(x)\| \pi(x)) + \KL(Q(\omega|x)\|\pi(\omega|x)),
\end{align}
where $Q(x)$ is the marginal distribution of $Q(\omega)a(x|\omega)$.
This implies that maximizing $\mathbb{E}[\log R]$ tends to reduce $\KL(Q(x)\|\pi(x))$.
Taking ``vanilla'' VI for example, given fixed $y$, a sample $\hat{x} \sim q$ plays the role of $\omega$, so $Q(\hat x)=q(\hat x)$, $R(\hat x) = p(\hat x, y)/q(\hat x)$ and $a(x|\hat x)=\delta_{\hat x}(x)$ specify an estimator-coupling pair for $p(x, y)$.
Then Equation \ref{equation3} reduces to the usual decomposition of VI that $\log p(x)=\mathcal{L}_{\scriptscriptstyle \text{ELBO}} + \KL(q(x) \| p(x \vert y))$ \citep{Blei2016VariationalIA}. For a more general VI, $Q$ is the sampling algorithm, $\mathbb{E}[\log R]$ gives the lower bound, and $Q\cdot a$ gives the augmented approximate posterior. A key observation is that operations on estimator-coupling pairs can be used to derive new variational objectives. An example is the Rao-Blackwellization operation~\citep{domke2019divide}:
\begin{lemma}
\label{lemma1}
Suppose that $R_0(\omega,\nu)$ and $a_0(x|\omega,\nu)$ are a valid estimator-coupling pair for $\gamma(x)$ under $Q_0(\omega,\nu)$. Then
\begin{align*}
R(\omega)&=\mathbb{E}_{Q_0(\nu|\omega)}R_0(\omega,\nu),\\
a(x|\omega)&=\frac{1}{R(\omega)}\mathbb{E}_{Q_0(\nu|\omega)}[R_0(\omega,\nu)a_0(x|\omega,\nu)]
\end{align*}
are a valid estimator-coupling pair for $\gamma(x)$ under $Q(\omega) = \int Q_0(\omega, \nu)d\nu$. Furthermore, $R$ has lower variance and gives a tighter bound than $R_0$, i.e., $\Var R(\omega) \leq \Var R_0(\omega, \nu)$ and $\E \left[ \log  R_0(\omega, \nu) \right] \leq \E \left[\log R(\omega)\right] \leq \log Z$.
Denote this operation by $(Q, R, a) = \textsc{Marginalize}(Q_0, R_0, a_0; \nu)$.
\end{lemma}
The \textsc{Marginalize} operation reduces the variance of an estimator by marginalizing some variables. It is an example of what is often called Rao-Blackwellization, where an estimator is replaced by its conditional expectation to reduce variance. This operation is the biggest difference between VMPF and VSMC.

\textbf{Variational Sequential Monte Carlo.} SMC~\citep{naesseth2019elements} is an algorithm that
constructs approximations for the sequence of target distributions $p(x_{1:t}|y_{1:t})$ using weighted samples.
In particular, $p(x_{1:t} | y_{1:t})$ is approximated by a weighted set of particles $(w_t^i, x^{t,i}_{1:t})_{i=1}^N$ as
\begin{align*}
  \hat\pi_t(x_{1:t}) = \sum_{i=1}^N \overline{w}_t^i\delta_{x_{1:t}^{t,i}}(x_{1:t}), \text{ where } \bar{w}^i_t = w_t^i/\Big(\textstyle \sum_{j=1}^N w_t^j\Big).
\end{align*}
We use a notation that explicitly distinguishes particles at different times, so $x^{t,i}_s$ is the $s$th entry of the $i$th particle in iteration $t$ of SMC, and $x^{t,i}_{1:t}$ is the entire particle. Later, we will use $x^{t,1:N}_{1:t}$ to denote the entire collection of $N$ particles.
The SMC procedure for sampling particles and computing weights is shown in Algorithm \ref{algorithm1}.
For each time step $t$ and particle $i$, an ancestor index $j$ is sampled (Line 6) and the corresponding particle is extended using the proposal distribution $r_t$ (Line 7), then weighted (Line 8) and assigned to become the new $i$th particle (Line 9).

From SMC, we get that $\hat{p}_{\scriptscriptstyle \text{SMC}}(y_{1:T})=\prod_{t=1}^T\frac{1}{N}\sum_{i=1}^Nw_t^i$ is an unbiased estimator for $p(y_{1:T})$.
Variational SMC uses $\mathcal{L}_{\scriptscriptstyle \text{VSMC}} =\mathbb{E}[\log\hat{p}_{\scriptscriptstyle \text{SMC}}(y_{1:T})]$ as a variational objective~\citep{naesseth2018variational,maddison2017filtering,le2017auto}. \citet{naesseth2018variational} show that 
\begin{align}
\label{eq:VSMC_bounds}
\mathcal{L}_{\scriptscriptstyle \text{VSMC}} \leq \mathcal{L}_{\sss{ELBO}}^{\sss{SMC}} \leq \log p(y_{1:T}),
\end{align}


where $\mathcal{L}_{\sss{ELBO}}^{\sss{SMC}}$ is the ELBO between the proposal $Q(x_{1:T})$ implied by SMC and the target $p(x_{1:T}|y_{1:T})$. VEM is used to simultaneously learn parameters of $p$ and
adjust the proposal distributions $r_t$ by maximizing $\mathcal{L}_{\sss{VSMC}}$. In practice, biased gradient estimates are most commonly used: the categorical ancestor variables (Line 6) cannot be reparameterized, and other estimators give problematically high variance~\citep{naesseth2018variational,maddison2017filtering,le2017auto}.
A recent paper uses ensemble particle transformations in place of resampling to obtain a fully differentiable filter (DPF)~\citep{corenflos2021differentiable}. 
While this introduces some bias into the likelihood estimator and so does not give a provable lower bound on the log-likelihood, unbiased gradients can be computed using reparameterization methods. We will compare to DPF in Section~\ref{Experiments}.

\begin{figure*}[t]
  \centering
  \begin{minipage}{.48\linewidth}
    \begin{algorithm}[H]
		\caption{Sequential Monte Carlo} 
		\label{algorithm1}
		\begin{algorithmic}[1]
			\Require $p(x_{1:T},y_{1:T})$, $y_{1:T}$, $\{r_t(x_t|x_{t-1})\}$,  $N$
                        \State Sample $x^{1,i}_1 \sim r_1(x_1)$ for all $i$
                        \State Set $w_1^i = \frac{f_1(x_1^{1,i})g(y_1|x_1^{1,i})}{r_1(x_1^{1,i})}$ for all $i$
			\For {$t=2,\ldots,T$}
                        \For {$i=1,\ldots,N$}
                        \State Set $\overline{w}_{t-1}^{j} = w_{t-1}^j / \sum_{k=1}^N w_{t\!-1}^k$ for all $j$
                        \State Sample $j  \sim \text{Categorical}\left(\overline{w}_{t-1}^{1:N}\right)$
                        \State Sample $x^{t,i}_t \sim r_t(x_t | x^{t-1,j}_{t-1})$
			\State Set $w_t^i=
                        \frac{f(x_t^{t,i}| x_{t-1}^{t-1,j})g(y_t|x_t^{t,i})}{r_t(x_t^{t,i} | x_{t-1}^{t-1,j})}$
                        \State Set $x^{t,i}_{1:t} = (x^{t-1,j}_{1:t-1}, x^{t,i}_t)$
                        \EndFor
			\EndFor
		\end{algorithmic}
	\end{algorithm}
  \end{minipage}
  \hfill
    \begin{minipage}{.48\linewidth}
    \begin{algorithm}[H]
		\caption{Marginal Particle Filter} 
		\label{algorithm2}
		\begin{algorithmic}[1]
			\Require $p(x_{1:T},y_{1:T})$, $y_{1:T}$, $\{r_t(x_t|x_{t-1})\}$,  $N$
                        \State Sample $x^{i}_1 \sim r_1(x_1)$ for all $i$
                        \State Set $v_1^i = \frac{f_1(x_1^{i})g(y_1|x_1^{i})}{r_1(x_1^{i})}$ for all $i$
			\For {$t=2,\ldots,T$}
                        \For {$i=1,\ldots,N$}
                        \State Set $\overline{v}_{t-1}^{j} = v_{t-1}^j / \sum_{k=1}^N v_{t\!-1}^k$ for all $j$
                        \State Sample $j  \sim \text{Categorical}\left(\overline{v}_{t-1}^{1:N}\right)$
                        \State Sample $x^{i}_t \sim r_t(x_t | x^{j}_{t-1})$
			\State Set $v_t^i= \frac{\sum_{j=1}^Nv_{t-1}^jf(x_t^i|x_{t-1}^j)g(y_t|x_t^i)}{\sum_{j=1}^N v_{t-1}^jr_t(x_t^i | x_{t-1}^j)}$
                        \EndFor
			\EndFor
		\end{algorithmic}
	\end{algorithm}
  \end{minipage}
\end{figure*}

\section{Couplings and Sequential Monte Carlo}
\label{section3}
In this section we show how SMC can be derived with operations on estimator-coupling pairs. This gives a straightforward and novel proof of unbiasedness, which is the key property that guarantees a lower bound of $\log p(y_{1:T})$ when used with VI. It will also be the basis of our MPF analysis.

We first give Lemma \ref{lemma2}, which replicates an estimator-coupling pair $N$ times.
\begin{lemma}
	\label{lemma2}
	Suppose that $R_0(\omega,\nu)$ and $a_0(x|\omega,\nu)$ are a valid estimator-coupling pair for $\gamma(x)$ under $Q_0(\omega,\nu)=Q_0(\omega)Q_0(\nu|\omega)$. Then
\begin{align*}
R(\omega,\nu_1,...,\nu_N)&=\frac{1}{N}\sum_{i=1}^NR_0(\omega,\nu_i),\\
a(x|\omega,\nu_1,...,\nu_N)&=\frac{\sum_{i=1}^NR_0(\omega,\nu_i)a_0(x|\omega,\nu_i)}{\sum_{i=1}^NR_0(\omega,\nu_i)}
\end{align*}
are a valid estimator-coupling pair for $\gamma(x)$ under $Q(\omega,\nu_1,...,\nu_N)=Q_0(\omega)\prod_{i=1}^NQ_0(\nu_i|\omega)$. Denote this operation by $(Q, R, a) = \textsc{Replicate}(Q_0, R_0, a_0; \nu, N)$.
\end{lemma}
 Lemma \ref{lemma2} is used at each step of SMC to get $N$ independent particles. It is a slight generalization of the ``IID Mean'' method in~\citep{domke2019divide}. 

All operations so far only work for a fixed target distribution. In SMC, we also need some operation to change the target distribution. For instance, at time~$t$, the target distribution should be extended from $p(x_{1:t-1},y_{1:t-1})$ to $p(x_{1:t},y_{1:t})$. Lemma \ref{lemma3} describes how to extend the target distribution.

\begin{lemma}
	\label{lemma3}
	Suppose that $R_0(\omega)$ and $a_0(x|\omega)$ are a valid estimator-coupling pair for $\gamma(x)$ under $Q_0(\omega)$, and $\gamma'(x,x')$ is an unnormalized distribution on the augmented space of $(x,x')$. Also suppose that we have a proposal distribution $r(x'|x)$ such that if $r(x'|x)=0$ then $\gamma'(x,x')/\gamma(x)=0$. Then
	\begin{align*}
		R(\omega,\hat x, \hat x')&=R_0(\omega)\frac{\gamma'(\hat x,\hat x')/\gamma(\hat x)}{r(\hat x'|\hat x)},\\
		a(x,x'|\omega, \hat x, \hat x')&=\delta_{(\hat x, \hat x')}(x,x')
	\end{align*}
	are a valid estimator-coupling pair for $\gamma'(x,x')$ under $Q(\omega, \hat x, \hat x')=Q_0(\omega)a_0(\hat x|\omega)r(\hat x'|\hat x)$. Denote this operation by $(Q, R, a) = \textsc{ExtendTarget}(Q_0, R_0, a_0; \gamma, \gamma', r)$.

If instead $a(x' | \omega, \hat x, \hat x') = \delta_{\hat x'}(x')$, then $(R,a)$ is still a valid estimator-coupling pair for $\gamma'(x') = \int \gamma'( x, x')dx$ under $Q$. Denote this by $(Q, R, a) = \textsc{ChangeTarget}(Q_0, R_0, a_0; \gamma, \gamma', r)$.
\end{lemma}

We now use these results to derive SMC with estimator-coupling pairs.
\begin{theorem}
\label{theoremSMC}
For the SSM in Equation~\eqref{SSM}, given fixed $y_{1:T}$
\begin{align}
  \label{eq:R_SMC}
  R\big(x_{1:1}^{1,1:N}, \ldots, x_{1:T}^{T,1:N}\big)&=\prod_{t=1}^T\frac{1}{N}\sum_{i=1}^N w_t^i,\\
  \label{eq:a_SMC}
  a\big(x_{1:T} \mid x_{1:1}^{1,1:N}, \ldots, x_{1:T}^{T,1:N}\big)& = \sum_{i=1}^N \overline{w}_T^i \delta_{x_{1:T}^{T,i}}(x_{1:T})
\end{align}
form an estimator-coupling pair for $p(x_{1:T}, y_{1:T})$ under the sampling distribution of Algorithm~\ref{algorithm1} with weights $w_t^i$ as given in Lines 2 and 8 and $\overline{w}_t^i = w_t^i/\big(\sum_{j=1}^N w_t^j\big)$. Thus, for any test function $h$,
\begin{align}
  &\mathbb{E} \left[ \left(\prod_{t=1}^T \frac{1}{N}\sum_{i=1}^N w_t^i\right) \cdot \sum_{i=1}^N \overline{w}_T^i h\big(x^{T,i}_{1:T}\big) \right]\notag\\
  = &p(y_{1:T}) \cdot \mathbb{E}_{p(x_{1:T} \vert y_{1:T})} \Big[ h(x_{1:T}) \Big].
  \label{eq:unbiased}
\end{align}
\end{theorem}

While the unbiasedness conclusion of Equation~\eqref{eq:unbiased} is well-known~\cite[e.g.,  ]{naesseth2019elements,maddison2017filtering}, we give a novel proof that breaks SMC into small operations. This proof strategy will form the basis for understanding MPF as applying Rao-Blackwellization operations within SMC.

\begin{proof}[(Proof sketch)]
To start, observe that the estimator $R_1(\hat x_1)=\frac{f(\hat x_1)g(y_1|\hat x_1)}{r_1(\hat x_1)}=\frac{p(\hat x_1,y_1)}{r_1(\hat x_1)}$ and the coupling $a_1(x_1 \vert \hat x_1)=\delta_{\hat x_1}(x_1)$ are valid for $p(x_1,y_1)$ under $Q_1(\hat x_1)=r_1(\hat x_1)$.
Now, for $t>1$, define 
\begin{align}
  &(Q_t, R_t, a_t) = \textsc{ExtendTarget}\Big(Q_{t-1}^N, R_{t-1}^N, a_{t-1}^N;\notag\\
   &\quad \quad p(x_{1:t-1}, y_{1:t-1}), p(x_{1:t}, y_{1:t}), r_t(x_t | x_{t-1}) \Big),
  \label{eq:SMC_extend_target}
\end{align}
where for all $t$,
\begin{align*}
  (Q_t^N, R_t^N, a_t^N) &= \textsc{Replicate}\left(Q_t, R_t, a_t; x^t_{1:t}, N\right).
\end{align*}

Mechanically applying these transformations, Lemmas \ref{lemma3} and \ref{lemma2} yield that the functions $R_T^N$ and $a_T^N$ match Equations~\eqref{eq:R_SMC} and \eqref{eq:a_SMC} and
\begin{align}
  &Q_T^N\big( x_{1:1}^{1,1:N}, \ldots, x_{1:T}^{T,1:N}\big) =\notag\\
  &\prod_{i=1}^N\Bigg[r_1(x_{1}^{1,i})\prod_{t=2}^T \sum_{j=1}^N \overline{w}_{t-1}^j \delta_{x^{t-1,j}_{1:t-1}}(x_{1:t-1}^{t,i})r_t(x_{t}^{t,i}|x_{t-1}^{t,i})\Bigg],
  \label{eq:Q_SMC}
\end{align}
which matches the SMC sampling distribution. The claimed result then follows immediately from the fact that $R^N_T$ and $a^N_T$ are a valid estimator-coupling pair for $p(x_{1:T},y_{1:T})$ under $Q_T^N$.
\end{proof}

Details appear in the supplement.
This proof uses induction on a sequence of estimators with operations that match the steps of the algorithm, and may be easier to understand than proofs that reason about the full expectation and require manipulating complex expressions. In addition, using operations on estimator-coupling pairs, it is possible to \textit{implement} estimators by transforming simple estimators in a way that exactly matches their derivation, which is of interest in probabilistic programming \citep{van2018introduction}.
\citet{douc2008limit} give a related framework to show consistency and asymptotic normality for SMC using operations on weighted particle systems that preserve those properties. \citet{stites2021learning} also derive SMC by operations on proper weighting \citep{liu2001monte, NaessethLS15}.


Since SMC can be derived by estimator-coupling pairs, we directly have that $\mathcal{L}_{\scriptscriptstyle \text{VSMC}} \leq \mathcal{L}_{\sss{ELBO}}^{\sss{SMC}} \leq \log p(y_{1:T})$ as in Equation~\eqref{eq:VSMC_bounds}. This result also quantifies the gap $\mathcal{L}_{\sss{ELBO}}^{\sss{SMC}} - \mathcal{L}_{\sss{VSMC}}$ as a conditional KL divergence. 

\section{Variational Marginal Particle Filter}

We now show that the marginal particle filter (MPF) of \citet{klaas2012toward} can also be derived with estimator-coupling pairs, which proves it is unbiased and suitable for use within VI. It uses \textsc{Marginalize} operations not present in SMC, which reduce variance and make VI bounds tighter ``locally''. Unlike SMC, it uses mixture proposals that can be reparameterized.

In Theorem~\ref{theoremSMC}, we see that SMC is not fully reparameterizable because of the Dirac distributions in $Q_T^N$ (Equation~\eqref{eq:Q_SMC}), which correspond to the sampling and copying operations in Lines 6 and 9 in Algorithm~\ref{algorithm1}. The non-reparameterizable variables $x_{t,1:t-1}^{t,i}$ are exactly the first $t-1$ entries of each particle. Our general idea is to marginalize these variables using Lemma \ref{lemma1} to get MPF.

\textbf{MPF and Couplings.}
The MPF algorithm is shown in Algorithm~\ref{algorithm2}. Instead of $p(x_{1:t}, y_{1:t})$, it targets the sequence of marginal distributions $p(x_{t}, y_{1:t})$ for all $t$. The procedure is very similar to SMC, \emph{but with different weights} for $t > 1$. In MPF, the $i$th marginal particle at time $t$ is denoted as $x_t^i$. Using this notation for both algorithms to facilitate comparison, the weights are:
\begin{align}
&\text{SMC:}\ w_t^i = \frac{f(x_t^i|x_{t-1}^j)g(y_t|x_t^i)}{r_t(x_t^i | x_{t-1}^j)},\  j \sim \text{\small Categorical}(\cdot), \notag\\
&\text{MPF:}\ v_t^i = \frac{\sum_{j=1}^Nv_{t-1}^jf(x_t^i|x_{t-1}^j)g(y_t|x_t^i)}{\sum_{j=1}^N v_{t-1}^jr_t(x_t^i | x_{t-1}^j)}.\notag
\end{align}
MPF can be obtained from SMC by two steps: (1)~drop the first $t-1$ variables $x_{1:t-1}$ from the target distribution and all particles to target $p(x_t, y_{1:t})$ instead of $p(x_{1:t}, y_{1:t})$, (2)~Rao-Blackwellize the ancestor index $j$ from the sampling distribution using the \textsc{Marginalize} operation in each step of SMC. 
Formally, we have:

\begin{theorem}
\label{theoremMPF}
For the SSM in Equation~\eqref{SSM}, given fixed $y_{1:T}$
\begin{align}
  \label{eq:R_MPF}
  R\big(x_1^{1:N}, \ldots, x_T^{1:N} \big)&=\prod_{t=1}^T\frac{1}{N}\sum_{i=1}^N v_t^i,\\
  \label{eq:a_MPF}
  a(x_T \mid x_1^{1:N}, \ldots, x_T^{1:N})&=\sum_{i=1}^N\overline{v}_T^i \, \delta_{x_{T}^{i}}(x_{T})
\end{align}
form an estimator-coupling pair for $p(x_T, y_{1:T})$ under the sampling distribution of Algorithm~\ref{algorithm2} with weights
$v_t^i$ as specified in Lines 2 and 8 and $\overline{v}_t^i = v_t^i/\big(\sum_{j=1}^N v_t^j\big)$. Thus, for any test function $h$,
\begin{align}
  &\mathbb{E} \left[ \left(\prod_{t=1}^T \frac{1}{N}\sum_{i=1}^N v_t^i\right) \cdot \sum_{i=1}^N \overline{v}_T^i\, h\big(x^{i}_{T}\big) \right]\notag\\
  =& p(y_{1:T}) \cdot \mathbb{E}_{p(x_{T} \vert y_{1:T})} \Big[ h(x_{T}) \Big].
  \label{eq:unbiased_MPF}
\end{align}
\end{theorem}
We are not aware of an existing proof of unbiasedness for MPF, though unbiasedness of the normalizing constant estimate (i.e., $h \equiv 1$ in Equation~\eqref{eq:unbiased_MPF}) can be derived from tensor Monte Carlo (TMC)~\citep{Aitchison19} (See Supplement 3) or a recent result on auxiliary particle filters~\citep{branchini2020optimized}. Our proof again shows that MPF is obtained by operations on estimator-coupling pairs.
\begin{proof}[(Proof sketch)]
The proof is very similar to the proof of Theorem~\ref{theoremSMC}, except the \textsc{ExtendTarget} operation in Equation~\eqref{eq:SMC_extend_target} for $t > 1$ is replaced by the following two operations:
\begin{align*}
(Q'_t, R'_t, a'_t) &= \textsc{ChangeTarget}\Big(Q_{t-1}^N, R_{t-1}^N, a_{t-1}^N; \\
&p(x_{t-1}, y_{1:t-1}), p(x_{t-1},x_t, y_{1:t}), r_t(x_t | x_{t-1}) \Big), \\
  (Q_t, R_t, a_t) &= \textsc{Marginalize}\Big(Q'_{t}, R'_{t}, a'_{t}; \hat x_{t-1} \Big).
\end{align*}
Recall that \textsc{ChangeTarget} is the same as \textsc{ExtendTarget} (Lemma~\ref{lemma3}), but drops $x_{t-1}$ from the target distribution. The \textsc{Marginalize} operation then \emph{marginalizes the corresponding variable} from the ``internal state'' of the estimator.
After mechanically applying the transformations of Lemma \ref{lemma3}, Lemma \ref{lemma1}, and Lemma~\ref{lemma2}, we get that the functions $R_T^N$ and $a_T^N$ match Equations~\eqref{eq:R_MPF} and \eqref{eq:a_MPF} and
\begin{align}
  &Q_T^N\big(x_1^{1:N}, \ldots, x_T^{1:N}\big)\notag \\
  =&\prod_{i=1}^N \Bigg[r_1(x_{1}^{i})\prod_{t=2}^T \sum_{j=1}^N \overline{v}_{t-1}^j r_t(x_{t}^{i}|x_{t-1}^{j})\Bigg],
  \label{eq:Q_MPF}
\end{align}
which matches the MPF sampling distribution. 
\end{proof}

In the proof sketch, we can see that the estimator $R_t$ following $\textsc{Marginalize}(\cdot)$ has lower (or the same) variance and gives a tighter (or the same) bound as $R_t'$, by Lemma~\ref{lemma1}. The same reasoning implies that using MPF weights is never worse than using the SMC weights ``locally'': in iteration~$t$, when targeting $p(x_{t}, y_{1:t})$, variance is never higher when using the MPF weight calculation in place of the SMC weight calculation, given the weights from iteration $t-1$. This does not necessarily imply the full MPF estimator has lower variance than SMC, but empirical evidence points to it having lower variance~\citep{klaas2012toward}.


The time complexity of MPF is $O(N^2T)$ compared to $O(NT)$ for SMC. In Line 8 of the MPF algorithm, the density $f(x_t^i | x_{t-1}^j)$ must be computed $N^2$ times in total vs. $N$ times in total in Line 8 of SMC. But there are only $N$ different conditional distributions: the distributions $f(\cdot | x_{t-1}^j)$ for each particle at the previous time-step. The density calculations can be split into two parts: (1) $O(N)$ pre-processing for each conditional distribution, and (2) evaluating the density $N^2$ times. For many models, the $O(N)$ pre-processing takes a significant fraction of the time.
For example, in our DMM experiments, pre-processing requires neural networks computation and density evaluation only elementary tensor operations, and the times of VMPF and VSMC are indistinguishable for $N \leq 16$.
%
Section 2.4 of \citet{Aitchison19} gives a similar argument.

\textbf{Variational MPF.}
Let $\pMPF(y_{1:T}) = \prod_{t=1}^N \frac{1}{N} \sum_{i=1}^N v_t^i$ be the unbiased estimator of $p(y_{1:T})$ from Equation~\eqref{eq:R_MPF}. We propose the variational objective
\begin{align}
\label{equationvrbsmc}
\mathcal{L}_{\sss{VMPF}}(\phi,\theta)=\mathbb{E}\left[\log\pMPF(y_{1:T};\phi,\theta)\right],
\end{align}
where $\phi$ stands for the parameter of proposal distributions $r_t$ and $\theta$ stands for the model parameters. By properties of estimator-coupling pairs, we immediately have that $\mathcal{L}_{\sss{VMPF}}(\phi,\theta)\le \mathcal{L}_{\sss{ELBO}}^{\sss{MPF}}(\phi,\theta)\le \log p(y_{1:T})$,
where $\mathcal{L}_{\sss{ELBO}}^{\sss{MPF}}(\phi,\theta)$ is the ELBO between the marginal proposal $Q(x_{T}; \phi)$ implied by the MPF procedure and the target distribution $p(x_{T} | y_{1:T}; \theta)$, and we can also quantify the gap $\mathcal{L}_{\sss{ELBO}}^{\sss{MPF}} - \mathcal{L}_{\sss{VMPF}}$ with Equation~\eqref{equation3}.
To compute and optimize this objective, we use Monte Carlo estimates for the value and gradients. We have two approaches to estimate the gradients.

\textbf{Biased gradients with categorical sampling.}
The first approach follows VSMC \citep{naesseth2018variational}. We assume that each proposal distribution $r_t$ is reparameterizable
and compute gradients of Equation (\ref{equationvrbsmc}) as $\nabla\mathcal{L}_{\scriptscriptstyle \text{VMPF}}(\phi,\theta) = \mathbb{E}[\nabla \log \hat{p}_{\scriptscriptstyle \text{MPF}}(y_{1:T};\phi,\theta)] + g_{\text{score}}$,
where the first term uses the reparameterization trick~\citep{kingma2013auto, rezende2014stochastic}, but ignores gradient paths for probabilities of categorical variables, which cannot be reparameterized; and $g_{\text{score}}$ is a score function term to handle the categorical sampling~\citep{naesseth2018variational}. As with VSMC, we observe that estimates of $g_{\text{score}}$ have very high variance and lead to slow convergence. For this approach, we simply drop $g_{\text{score}}$ and estimate the first term, which is a biased gradient estimator. We call this method VMPF with biased gradient (VMPF-BG), which does not have any limitation on the proposals. 

\textbf{Unbiased gradients with implicit reparameterization.}
A biased gradient estimator can lead to suboptimal inference~\citep{corenflos2021differentiable}. A very appealing property of Algorithm \ref{algorithm2} is that, as a result of Rao-Blackwellization, the variables $j$ and $x^j_{t-1}$ are not used in weight computations, so Lines 6 and 7 can be combined conceptually into a single draw $x_t^i \sim \sum_{j=1}^N \overline{v}_{t-1}^j r_t(x_{t} | x_{t-1}^j)$ from a mixture distribution. This is evident from Equation~\eqref{eq:Q_MPF}, which includes only the (continuous) mixture density $\sum_{j=1}^N \overline{v}_{t-1}^j r_t(x_t^i | x_{t-1}^j)$ for $x_t^i$. It is therefore possible to reparameterize $x_t^i$ by sampling from the mixture and then using implicit reparameterization gradients~\citep{graves2016stochastic,figurnov2018implicit}. We then have the fully reparameterized gradient $\nabla\mathcal{L}_{\scriptscriptstyle \text{VMPF}}(\phi,\theta) = \mathbb{E}[\nabla \log \hat{p}_{\scriptscriptstyle \text{MPF}}(y_{1:T}; \phi, \theta)]$,
and can form an unbiased estimate by drawing samples and backpropagating through mixtures.
This is currently possible for any proposal that is a product distribution, or full-rank Gaussians.
We call this method VMPF with unbiased gradients (VMPF-UG). It is a fully reparameterized gradient estimate for a variational filtering objective that lower bounds the log-likelihood for suitable proposal distributions.


\section{Related Work}
\label{sec:related}
There is significant previous work on improving VI approximations. One direction enriches the variational family directly, for example, with normalizing flows \citep{papamakarios2019normalizing,RezendeM15}, copulas \citep{TranBA15,HanLDC16,HirtDD19}, or mixture distributions~\citep{miller2017variational}. Another direction increases expressiveness by introducing auxiliary variables: this work includes hierarchical variational models \citep{RanganathTB16}, VI with Markov chain Monte Carlo \citep{SalimansKW15, CateriniDS18}, variational Gaussian processes \citep{TranRB15}, and importance-weighted VI (IWVI)~\citep{BurdaGS15,domke2018importance}. Estimator-coupling pairs generalize IWVI and include other variance reduction techniques, such as stratified sampling~\citep{domke2019divide}. For SSMs, three papers independently proposed to use SMC as an unbiased estimator to generalize IWVI in another direction~\citep{naesseth2018variational,maddison2017filtering,le2017auto}. This work builds on the above two ideas.

One limitation of prior SMC variational objectives is that they are not fully differentiable due to resampling steps. \citet{Moretti2019ParticleSV} use the concrete distribution \citep{MaddisonMT17,JangGP17} to approximate the resampling step, but they focus on the signal-to-noise ratio problem \citep{rainforth2018tighter} and do not mention any performance improvement due to differentiability. A series of works employ differentiable neural networks to approximate the resampling function \citep{KarkusHL18,JonschkowskiRB18,ZhuTowards,MaKHL20,MaKHLY20}. However, none produces fully differentiable SMC~\citep{corenflos2021differentiable}. \citet{corenflos2021differentiable} use optimal transport to learn an ensemble transform to replace resampling, leading to the first fully differentiable particle filter in the literature. Their likelihood estimator is asymptotically consistent, but biased, so does not give a provable lower bound of the log-likelihood. 

Another interesting line of previous work is independent particle filters (IPF) \citep{lin2005independent}, which improve SMC with multiple permutations of ancestor variables. With the setting of complete matching, IPF becomes tensor Monte Carlo (TMC) \citep{Aitchison19} for SSMs.
We outline an alternate viewpoint of MPF as TMC with specific mixture proposals in Section 3 of the supplement.

Recently, \citet{campbell2021online} propose an online VI which outperforms several online filtering methods on SSM by using a Bellman-type recursion similar to those used in reinforcement learning~\citep{Sutton2005ReinforcementLA}. 

\section{Experiments}
\label{Experiments}
We conduct experiments on linear Gaussian SSMs, stochastic volatility models, and deep Markov models (DMMs) \citep{KrishnanSS17} and compare lower bounds obtained by \VSIS\footnote{This uses $q(x_{1:T}) = r_1(x_1)\prod_{t=2}^T r_t(x_t | x_{t-1})$ as proposal and is equivalent to VSMC without resampling.}, VSMC with biased gradients~\citep{naesseth2018variational}, tensor Monte Carlo (TMC)~\citep{Aitchison19} (factorized), differentiable particle filter (DPF)~\citep{corenflos2021differentiable} (evaluated by SMC), VMPF-BG, and VMPF-UG.
We implement all algorithms in TensorFlow with TensorFlow Probability \citep{tensorflow2015-whitepaper, TFD} and train with the Adam optimizer \citep{KingmaB14}. The code to replicate the experiments can be found at \href{https://github.com/lll6924/VMPF}{https://github.com/lll6924/VMPF}.

\paragraph{Linear Gaussian State Space Models}
We first test with linear Gaussian models, for which the exact log-likelihood can be computed by the Kalman filter (KF)~\citep{harvey1990forecasting}. The model is $x_t=Ax_{t-1}+v_t,\ y_t=Cx_t+e_t,$
where $v_t\sim \mathcal{N}(0,Q)$, $e_t\sim \mathcal{N}(0,R)$, and $x_1\sim \mathcal{N}(0,I)$. We follow \citet{naesseth2018variational} and set $T=10$, $(A)_{ij}=\alpha^{|i-j|+1}$ for $\alpha=0.42$, $Q=I$ and $R=I$. There are two settings for $C$: ``sparse'' $C$ has diagonal entries $1$ and other entries $0$; ``dense'' $C$ has $C_{ij}\sim \mathcal{N}(0,I)$ for all $i, j$. We vary $d_x=\dim(x_t)$,  $d_y=\dim(y_t)$, and whether $C$ is sparse or dense. The same model $p$ is used to generate data and during inference. We choose the proposal distributions $r_t(x_t|x_{t-1};\phi)=\mathcal{N}(x_t|\mu_t+\diag(\beta_t)Ax_{t-1},\diag(\sigma_t^2))$ with $\phi=(\mu_t,\beta_t,\sigma_t^2)_{t=1}^T$ and maximize each objective with respect to $\phi$. In all settings, we first train for 10K iterations with learning rate $0.01$, then another 10K iterations with learning rate $0.001$. Further lowering the learning rate has little effect.
\begin{figure*}[t]
\includegraphics[width=0.99\textwidth]{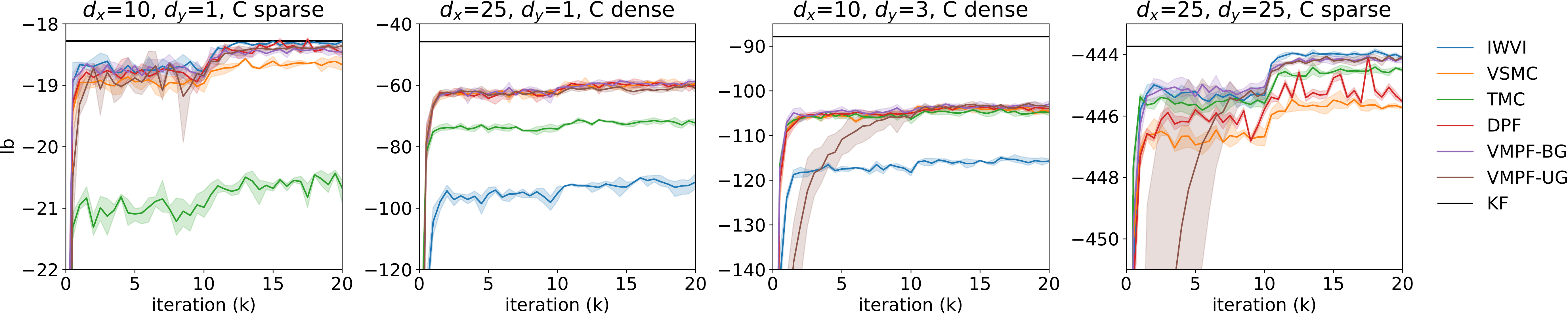}	
\caption{$\mathbb{E}[\log\hat{p}(y_{1:T})]$ as a function of iterations for IWVI, VSMC, DPF, VMPF-BG and VMPF-UG with $N=4$ under four settings for the linear Gaussian SSM. Black line: true log-likelihood. }
\label{ConvergenceFigure}
\vspace{-1pt}
\end{figure*}

We first examine convergence of each algorithm with $N=4$. Figure \ref{ConvergenceFigure} shows the lower bound of each method during training. 
For sparse $C$, the VMPF bounds are substantially higher than VSMC, but \VSIS is highest. The bound of DPF is between VSMC and VMPF in this case. 
In contrast, for dense $C$, \VSIS is much worse, while the final bounds of VSMC, DPF and VMPF are similar.
Overall, VMPF-BG is never worse than VSMC, TMC or DPF in terms of final bounds or convergence speed, and gives significantly higher bounds for sparse $C$. 
The convergence of VMPF-UG is slow,\footnote{For $d_x=25$, $d_y=1$, dense $C$ (second panel), 20K iterations were not enough to train VMPF-UG, so we initialiazed it with the parameters from VMPF-BG.}  and the final bound is comparable to, but not higher than, VMPF-BG.

It is likely \VSIS performs well for sparse $C$ because the variational family $q(x_{1:T}; \phi)$ includes the true posterior. Because SMC ``greedily'' resamples particles with high probability under $p(x_t| y_{1:t})$ (using only the first $t$ observations), it is counterproductive relative to a very accurate model of $p(x_t | y_{1:T})$ (conditioned on \emph{all} observations); see the discussion of sharpness in~\citep{maddison2017filtering}.
The variational family does not include the posterior for dense $C$, and IWVI is much worse than the SMC-based methods. 
To further understand this, we reran the sparse $C$ experiment for $d_x=25$, $d_y=25$ after fixing $\beta_t=1$ to impoverish the variational family. The final bounds become \textminus 456.20, \textminus 453.32, and \textminus 451.85 for IWVI, VSMC, and VMPF, respectively, confirming that resampling can be harmful when $q(x_{1:T}; \phi)$ can already approximate the true posterior very well, but tends to be beneficial otherwise.

%
%

%
%
%
\begin{figure*}[t]
  \centering
      \includegraphics[width=.42\textwidth]{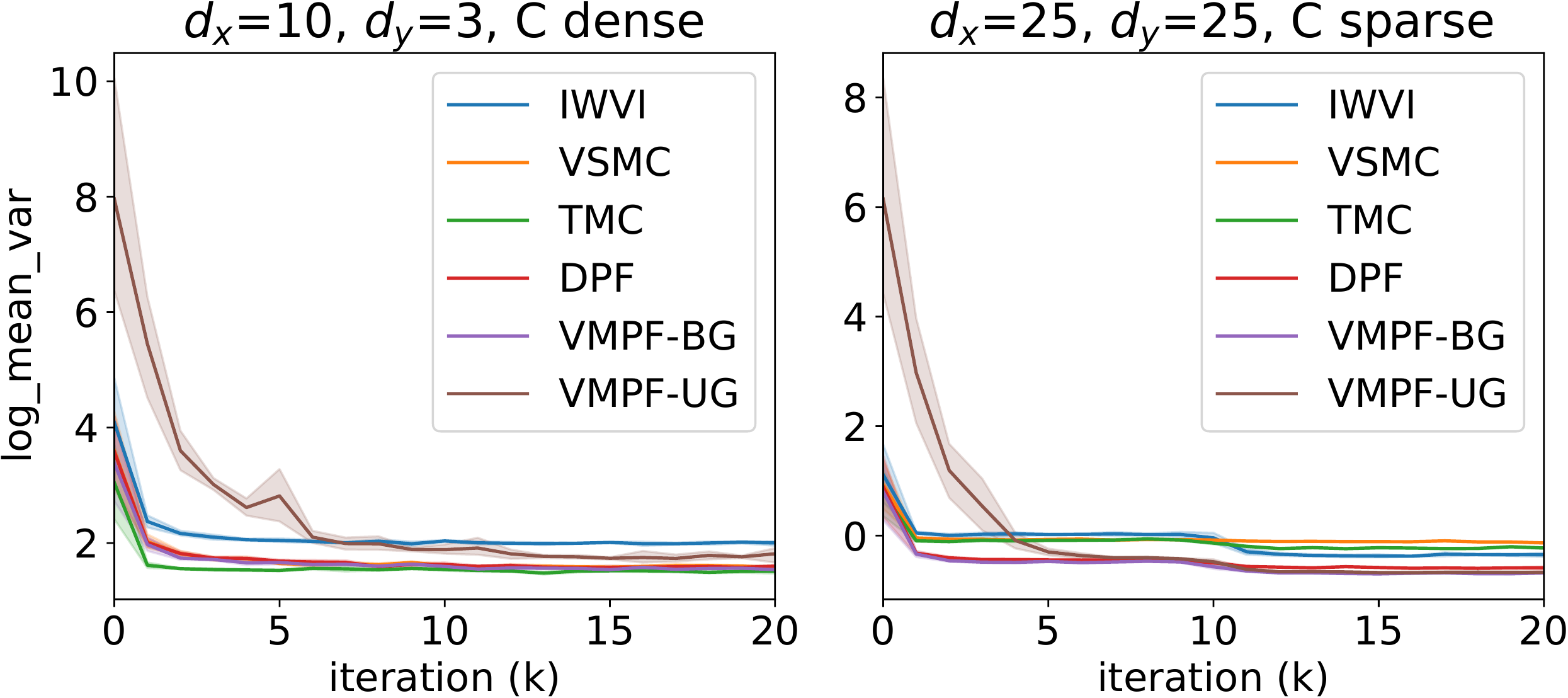}
  \hspace{1pt}
		\includegraphics[width=.56\textwidth]{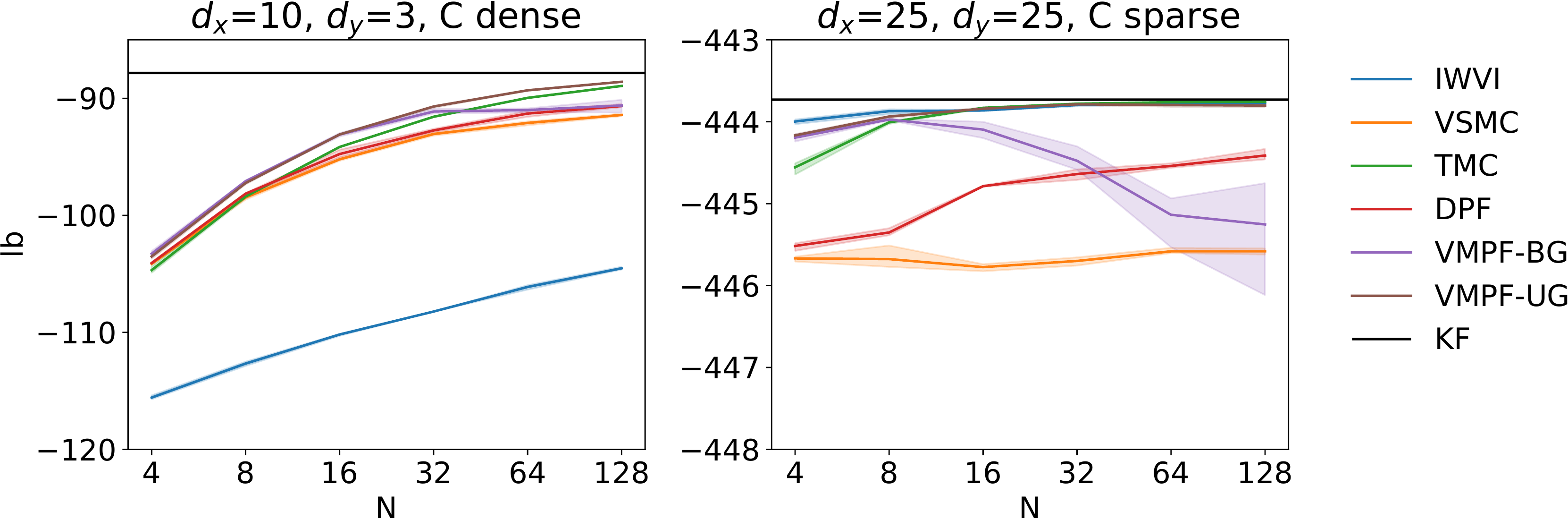}
    	\caption{Results for linear Gaussian SSMs. Left: logarithm of mean gradient variance vs. training iteration. Right: final lower bound vs. $N$. Experimental settings indicated in plots.}
    	\label{VarianceGradientsBiasGapFigure}
  \vspace{-1pt}
\end{figure*}
The slow convergence of VMPF-UG can be explained by gradient variance (Figure \ref{VarianceGradientsBiasGapFigure}, left). Both VSMC and VMPF with biased gradients have low variance and converge quickly. In contrast, VMPF-UG has high gradient variance and slower convergence, especially in early iterations, but variance reduces substantially when close to convergence. This suggests a strategy of using the biased gradient estimator at the beginning of optimization and then switching to the unbiased estimator.


Although we did not observe it for $N=4$, unbiased gradients can lead to tighter bounds upon convergence, especially for larger numbers of particles (Figure \ref{VarianceGradientsBiasGapFigure}, right).
For small $N$, VMPF-BG and VMPF-UG are similar, but as $N$ increases, the gap between the methods increases, which indicates that biased gradients are more of a problem. We conjecture that the bias of gradients for VMPF-BG will not go to zero as $N$ is increased, but the magnitude of the true gradient shrinks as $N \to \infty$, which leads to difficulty in training~\citep{le2017auto,rainforth2018tighter}.

 
\paragraph{Stochastic Volatility}
\label{SVexp}
The stochastic volatility model \citep{chib2009multivariate} is widely used for financial data. It is $x_t=\mu + \Phi(x_{t-1}-\mu) + v_t,\ y_t=\diag\left(\exp\left(x_t/2\right)\right)B e_t,$
where $v_t\sim\mathcal{N}(0,Q)$, $e_t\sim \mathcal{N}(0,I)$, and $x_1\sim\mathcal{N}(\mu,Q)$. The model parameters are $\theta=(\mu,\Phi,Q,B)$, where $\mu$ is a vector, $\Phi$ and $Q$ are diagonal matrices, and $B$ is either a diagonal or lower triangular matrix (with positive diagonal entries in both cases). We use VEM to learn $\theta$. Following \citet{naesseth2018variational}, we use the proposal $r_t(x_t|x_{t-1};\phi,\theta)\propto f(x_t|x_{t-1};\theta)\mathcal{N}(x_t;\mu_t,\Sigma_t)$
with parameters $\phi=(\mu_t,\Sigma_t)_{t=1}^T$ and $\Sigma_t$ diagonal.

\begin{table*}[t]
 \centering
\caption{Stochastic volatility model lower bounds for \VSIS, VSMC, DPF, VMPF-BG and VMPF-UG (higher is better). Mean and standard deviation of 3 runs are reported.}
\label{SVTable}
\begin{threeparttable}
	\begin{tabular}{cccccc}
&Method&$N=1$&$N=4$&$N=8$&$N=16$\\
\toprule
\multirow{6}*{Diagonal $B$}&\VSIS&\multirow{6}*{7216.09 (0.42)}&\textbf{7219.77} (0.17)&\textbf{7220.60} (0.51)&\textbf{7221.51} (0.10)\\
 &VSMC&&7200.57 (0.11)&7198.13 (0.30)&7197.70 (0.09)\\
 &TMC&&7193.21 (0.08)&7202.16 (0.07)&7208.67 (0.11)\\
  &DPF&&7209.42 (0.99)&7209.05 (1.13)&7210.22 (0.24)\\
 &VMPF-BG&&7205.29 (0.17)&7205.04 (0.16)&7205.90 (0.35)\\
  &VMPF-UG&&7208.57 (0.27)&7208.95 (0.28)&7206.35 (0.27)\\
 \midrule
  \multirow{6}*{Triangular $B$}&\VSIS&\multirow{6}*{8585.16 (1.00)}&\textbf{8590.87} (0.27)&\textbf{8593.19} (0.64)&\textbf{8595.78} (0.82)\\
  &VSMC&&8573.04 (0.23)&8572.58 (0.19)&8572.01 (0.08)\\
   &TMC&&8554.89 (0.18)&8570.56 (0.01)&8582.32 (0.10)\\
   &DPF&&8572.74 (2.97)&8574.32 (1.65)&8574.32 (0.32)\\
 &VMPF-BG&&8576.57 (0.61)&8578.53 (0.20)&8581.21 (0.39)\\
 &VMPF-UG&&8556.64 (3.63)&8543.40 (1.48)&8538.15 (5.36)\\
 \bottomrule
	\end{tabular}
\end{threeparttable}
\end{table*}

We model the exchange rates of 22 international currencies with respect to US dollars for 10 years (monthly from 4/2011 to 3/2021). The data can be downloaded from the US Federal Reserve System.\footnote{https://www.federalreserve.gov/releases/h10/current/}
Table \ref{SVTable} reports the optimized lower bound for different algorithms for $N\in\{4,8,16\}$. VMPF-BG always gives a higher bound than VSMC, and the \VSIS bound is always highest. TMC works well for $N=16$, but is much worse for smaller $N$. We also notice that DPF works slightly better than VSMC or VMPF on a simpler model with diagonal $B$, but worse than VMPF with triangular $B$. 
VMPF-UG is beneficial for diagonal $B$ with $N \in \{4,8\}$, but becomes increasingly harder to train with large $N$ and a larger model.
The IWVI performance is surprising and contrary to similar experiments in~\citep{naesseth2018variational}, but appears to be another case where the family of proposal distributions can already approximate the posterior very well: 
we also find the ELBO with ``vanilla'' VI ($N=1$) is higher than SMC-based methods with $N > 1$ in both cases.

\paragraph{Deep Markov Models}
\begin{table*}[t]
\centering
\caption{Test set nats per timestep for DMM trained with \VSIS, VSMC, DPF, VMPF-BG (higher is better) on four polyphonic music datasets. Mean and standard deviation of 3 runs are reported.}
\label{DMMTable}
\begin{threeparttable}
	\begin{tabular}{cccccc}
$N$&Method&Nottingham&JSB&MuseData&Piano-midi.de\\
\toprule
\multirow{5}*{4}&\VSIS& -3.86 (0.04)&-7.40 (0.01)& -8.19 (0.04)&-8.78 (0.01)\\
 &VSMC&-3.38 (0.03)& -7.16 (0.01)&-7.68 (0.02)&-8.39 (0.03)\\
  &TMC& -3.65 (0.02)&-7.51 (0.01)&-8.41 (0.02)&-9.00 (0.01)\\
  &DPF&-3.33 (0.02)&-7.14 (0.01)&-7.75 (0.01)&-8.45 (0.01)\\
 &VMPF-BG&\textbf{-3.29} (0.03)&\textbf{-7.04} (0.01)&\textbf{-7.67} (0.00)&\textbf{-8.35} (0.01)\\
 \midrule
  \multirow{5}*{8}&\VSIS&-3.82 (0.03)&-7.38 (0.01)&-8.16 (0.04)&-8.75 (0.01)\\
  &VSMC& -3.20 (0.02)&-6.96 (0.00)&-7.43 (0.01)&-8.20 (0.01)\\
  &TMC&-3.40 (0.02)&-7.24 (0.01)&-8.03 (0.01)&-8.64 (0.01)\\
  &DPF& -3.19 (0.00)& -6.95 (0.01)&-7.40 (0.01)&-8.31 (0.00)\\
 &VMPF-BG&\textbf{-3.09} (0.03)&\textbf{-6.80} (0.01)&\textbf{-7.39} (0.01)&\textbf{-8.12} (0.01)\\
 \midrule
  \multirow{5}*{16}&\VSIS& -3.84 (0.03)&-7.33 (0.02)&-8.16 (0.02)&-8.74 (0.02)\\
  &VSMC&-3.06 (0.02)&-6.81 (0.00)&-7.22 (0.02)&-8.03 (0.00)\\
  &TMC&-3.23 (0.02)&-6.98 (0.02)&-7.72 (0.00)&-8.37 (0.01)\\
  &DPF&-3.08 (0.02)&-6.78 (0.00)&-7.22 (0.01)&-8.18 (0.02)\\
 &VMPF-BG&\textbf{-2.96} (0.02)&\textbf{ -6.64} (0.01)& \textbf{-7.14} (0.01)& \textbf{-7.92} (0.01)\\
 \bottomrule
	\end{tabular}
\end{threeparttable}
\end{table*}

We evaluate the bounds of all methods for deep Markov models (DMMs) on four polyphonic music datasets: Nottingham, JSB, MuseData and Piano-midi.de \citep{Boulanger-LewandowskiBV12}. These are typically modeled with VRNNs, but the marginalization of ancestor variables prevents using VMPF for VRNNs (see Section~\ref{sec:limitations} and the supplement). The DMM model is
\begin{align}
	x_t&=\mu_{\theta}(x_{t-1}) + \mathrm{diag}(\exp(\sigma_{\theta}(x_{t-1})/2))v_t,\notag\\
	y_t&\sim \mathrm{Bernoulli}(\mathrm{sigmoid}(\eta_{\theta}(x_t))),\notag
\end{align}
where $v_t\sim \mathcal{N}(0,I)$, $x_0=0$, and $\mu_\theta$, $\sigma_\theta$, $\eta_\theta$ are neural networks. To approximate the posterior, we define the proposal distribution
\begin{align}
  r(x_t|x_{t-1},y_t;\phi)\propto &\mathcal{N}\big(x_t;\mu^x_{\phi}(x_{t-1}),\diag(\exp(\sigma^x_{\phi}(x_{t-1})))\big)\notag\\
  &\cdot \mathcal{N}\big(x_t;\mu^y_{\phi}(y_t),\diag(\exp(\sigma^y_{\phi}(y_t)))\big)\notag
\end{align}
where $\mu^x_\phi$, $\sigma^x_\phi$, $\mu^y_\phi$ and $\sigma^y_\phi$ are neural networks. Details can be found in supplement. Table \ref{DMMTable} shows the results of different methods. We see that in all cases VMPF-BG produces the best results. It is possible to train VMPF-UG in most settings here by initializing with VMPF-BG and/or using gradient clipping, but they lead to a slightly worse results than VMPF-BG.

\section{Limitations}
\label{sec:limitations}
\label{Limitation}
We are aware of several limitations or potential limitations.
First, for VMPF-UG, implicit reparameterization gradients for mixture distributions require the ability to compute conditional CDFs of each component distribution~\citep{graves2016stochastic,figurnov2018implicit}, which is straightforward for Gaussians or product distribution, but may be difficult in general. Current implementations support only product distributions~\citep{TFD}. At present, this limits the choice of proposals for VMPF-UG. There is no such limitation for VMPF-BG.

Second, although implicit reparameterization gives unbiased gradients for VMPF, we show that the variance remains high compared to VSMC and VMPF-BG,  which restricts applying VMPF-UG in some cases, especially to complex models. Future work can focus on reducing the variance.

Third, unlike VSMC, VMPF only works for the the marginal objective $p(x_t|y_{1:t})$, which restricts some applications, for example, to VRNNs~\citep{ChungKDGCB15}. See the supplement for discussion. 

\clearpage
\clearpage

\subsubsection*{Acknowledgements}
We thank Javier Burroni, Tomas Geffner and the anonymous reviewers for comments that greatly improved the manuscript. This material is based upon work supported by the National Science Foundation under Grant Nos. 1749854, 1908577 and 2045900.

\bibliographystyle{plainnat}
\bibliography{reference}
\balance


\clearpage
\appendix

\thispagestyle{empty}

\onecolumn \makesupplementtitle

\section{Proof of the Operations}
\subsection{Proof of Lemma 1}
\begin{proof}
	We have that 
	\begin{align}
		\int Q(\omega)R(\omega)a(x|\omega)d\omega\notag
		=&\int Q(\omega)R(\omega)\frac{1}{R(\omega)}\left(\int \frac{Q_0(\omega,\nu)}{Q(\omega)}R_0(\omega,\nu)a_0(x|\omega,\nu) d\nu\right)d\omega\notag\\
		=&\int\int Q_0(\omega,\nu)R_0(\omega,\nu)a_0(x|\omega,\nu)d\nu d\omega\notag\\
		=&\gamma(x).\notag
	\end{align}
	For the variance, we have
	\begin{align}
		\Var R_0(\omega,\nu)&=\Var\E [R_0(\omega,\nu)|\omega]+\E\Var [R_0(\omega,\nu)|\omega]\notag\\
		&\ge \Var\E [R_0(\omega,\nu)|\omega]\notag\\
		&=\Var R(\omega).\notag
	\end{align}
	For the lower bounds,
	\begin{align}
		\E_{\omega,\nu}[\log R_0(\omega,\nu)]&=\E_{\omega}\left[\E_{\nu|\omega}[\log R_0(\omega,\nu)]\right]\notag\\
		&\le \E_{\omega}\left[\log\E_{\nu|\omega}[ R_0(\omega,\nu)]\right]\notag\\
		&= \mathbb{E}_{\omega}[\log R(\omega)],\notag
	\end{align}
        and $\E_\omega[\log R(\omega)] \leq \log Z$ because we have shown that $R$ and $a$ are valid estimator-coupling pair for~$\gamma$.
\end{proof}
\subsection{Proof of Lemma 2}
\begin{proof}
	\begin{align}
\int&\dots\int Q(\omega,\nu_1,\ldots,\nu_N) R(\omega,\nu_1,\ldots,\nu_N) a(x|\omega,\nu_1,\ldots,\nu_N) d\omega\, d\nu_1 \ldots d\nu_N \notag\\
=&\int\dots\int Q_0(\omega) \left( \prod_{i=1}^N Q_0(\nu_i|\omega)\right)\!\left( \frac{1}{N}\sum_{i=1}^N R_0(\omega,\nu_i)\right) \frac{\sum_{i=1}^NR_0(\omega,\nu_i) a_0(x|\omega,\nu_i)}{\sum_{i=1}^NR_0(\omega,\nu_i)} d\omega\, d\nu_1 \ldots d\nu_N\notag\\
=&\frac{1}{N}\sum_{i=1}^N\int \cdots \int  Q_0(\omega) \prod_{j=1}^NQ_0(\nu_j|\omega) R_0(\omega,\nu_i) a_0(x|\omega,\nu_i) d\omega\, d\nu_1\ldots d\nu_N\notag\\
=&\frac{1}{N} \sum_{i=1}^N \int Q_0(\omega) Q_0(\nu_i|\omega) R_0(\omega,\nu_i) a_0(x|\omega,\nu_i) \,d\omega\, d\nu_i\notag\\
=& \gamma(x).\notag
        \end{align}
\end{proof}
\subsection{Proof of Lemma 3}
\begin{proof}
For $\textsc{ExtendTarget}$,
\begin{align}
	\int \int\int &Q(\omega,\hat x,\hat x')R(\omega,\hat x, \hat x')a(x, x'|\omega,\hat x, \hat x')\, d\hat x'\, d\hat x\, d\omega \notag\\
	=&\int \int\int Q_0(\omega)a_0(\hat x|\omega)r(\hat x'|\hat x)R_0(\omega)\frac{\gamma'(\hat x, \hat x')/\gamma(\hat x)}{r(\hat x'|\hat x)}\delta_{(\hat x,\hat x')}(x, x')\, d\hat x'\, d\hat x\, d\omega\notag\\
	=&\int\int\left(\int Q_0(\omega)a_0(\hat x|\omega)R_0(\omega)\, d\omega\right) \frac{\gamma'(\hat x, \hat x')}{\gamma(\hat x)} \delta_{(\hat x,\hat x')}(x, x')\, d \hat x'\, d \hat x\notag\\
	=&\int\int \gamma'(\hat x, \hat x') \delta_{(\hat x, \hat x')}(x,x')\, d\hat x'\, d\hat x \notag\\
	=& \gamma'(x, x').\notag
\end{align}
For $\textsc{ChangeTarget}$, proceed as above with $\delta_{\hat x'}(x')$ in place of $\delta_{(\hat x, \hat x')}(x,x')$. The steps are the same until the second to last line, which becomes
\begin{align}
\int\int \gamma'(\hat x, \hat x') \delta_{\hat x'}(x')\, d\hat x' \, d \hat x\notag 
= \int \gamma'(\hat x, x') d\hat x\notag
= \gamma'(x').\notag
\end{align}
\end{proof}

\section{Derivation of SMC and MPF}
\subsection{Proof of Theorem 1}
We first repeat relevant definitions from the algorithm and theorem statement. The weights are 
\begin{align}
	w_1^i=\frac{f(x_{1}^{1,i})g(y_1|x_{1}^{1,i})}{r_1(x_{1}^{1,i})}, \qquad
	w_t^i=\frac{f(x_{t}^{t,i}|x_{t-1}^{t,i})g(y_t|x_{t}^{t,i})}{r_t(x_{t}^{t,i}|x_{t-1}^{t,i})} \text{ for $t > 1$}.\notag
\end{align}
The normalized weights are $\overline{w}_t^i=w_t^i/\left(\sum_{j=1}^Nw_t^j\right)$.

We wish to show that, for all $t$,
\begin{align}
  Q_{t}^N(x_{1:1}^{1,1:N}, \ldots , x_{1:t}^{t,1:N}) &= \prod_{i=1}^N\left[ r_1(x_{1:1}^{1,i}) \prod_{\tau=2}^{t} \sum_{j=1}^N \overline{w}_{\tau-1}^j \delta_{x_{1:\tau-1}^{\tau-1,j}} (x_{1:\tau-1}^{\tau,i}) r_\tau(x_\tau^{\tau,i}|x_{\tau-1}^{\tau,i}) \right], \label{eq:Q_SMC_proof} \\
  R_{t}^N(x_{1:1}^{1,1:N},...,x_{1:t}^{t,1:N}) &= \prod_{\tau=1}^{t} \frac{1}{N} \sum_{i=1}^N  w_\tau^i, \label{eq:R_SMC_proof} \\
  a_{t}^N(\cdot| x_{1:1}^{1,1:N}, \ldots , x_{1:t}^{t,1:N}) &= \sum_{i=1}^N \overline{w}_t^i \delta_{x_{1:t}^{t,i}}(\cdot) \label{eq:a_SMC_proof} 
\end{align}
define an estimator-coupling pair for $p(x_{1:t}, y_{1:t})$.

We can check that these match the SMC algorithm at step $t$: that is, $Q_t^N$ is sampling distribution, $R_t^N$ is the likelihood estimator, and $a_t^N$ is approximation to $p(x_{1:t}|y_{1:t})$. So, after proving this, the conclusion of the theorem follows immediately.

We will show inductively that $(Q_t^N, R_t^N, a_t^N)$ are obtained by applying appropriate operations on estimator-coupling pairs. In particular, the procedure is 
\begin{align}
  (Q_t^N, R_t^N, a_t^N) &= \textsc{Replicate}\left(Q_t, R_t, a_t; x^t_{1:t}, N\right),\notag
\end{align}
where
\begin{align}
	Q_1(x_{1:1}^{1})&=r_1(x_{1}^{1}),\notag\\
	R_1(x_{1:1}^{1})&=\frac{p(x_{1}^{1},y_1)}{r_1(x_{1}^{1})}=\frac{f(x_{1}^{1})g(y_1|x_{1}^{1})}{r_1(x_{1}^{1})},\notag\\
	a_1(\cdot|x_{1:1}^{1})&=\delta_{x_{1:1}^{1}}(\cdot),\notag
\end{align}
and, for $t > 1$, 
\begin{align}
	(Q_t, R_t, a_t) = \textsc{ExtendTarget}\Big(Q_{t-1}^N, R_{t-1}^N, a_{t-1}^N; p(x_{1:t-1}, y_{1:t-1}), p(x_{1:t}, y_{1:t}), r_t(x_t | x_{t-1}) \Big). \notag
\end{align}

The proof is a mechanical application of these operations.

For the base case, it is immediate that $(Q_1, R_1, a_1)$ are an estimator-coupling pair for $p(x_1, y_1)$, and easy to verify that applying Lemma 2 yields $(Q_1^N, R_1^N, a_1^N)$ that match Equations~\eqref{eq:Q_SMC_proof}--\eqref{eq:a_SMC_proof} and give an estimator-coupling pair for $p(x_1, y_1)$.

For the induction step ($t > 1$), we first apply Lemma 3 to obtain $(Q_t, R_t, a_t)$ from $(Q_{t-1}^N, R_{t-1}^N, a_{t-1}^N)$ using the \textsc{ExtendTarget} operation. We get
\begin{align*}
Q_t(x_{1:1}^{1,1:N}, \ldots ,x_{1:t-1}^{t-1,1:N},x_{1:t}^t)
&= Q_{t-1}^N(x_{1:1}^{1,1:N}, \ldots, x_{1:t-1}^{t-1,1:N}) \\
   & \qquad \cdot a_{t-1}^N(x_{1:t-1}^t|x_{1:1}^{1,1:N}, \ldots ,x_{1:t-1}^{t-1,1:N}) \cdot r_t(x_t^t|x_{t-1}^t) \notag\\[2pt]
&= \prod_{i=1}^N \left[ r_1(x_{1:1}^{1,i}) \prod_{\tau=2}^{t-1} \sum_{j=1}^N \overline{w}_{\tau-1}^j \delta_{x_{1:\tau-1}^{\tau-1,j}}(x_{1:\tau-1}^{\tau,i}) r_\tau(x_\tau^{\tau,i} | x_{\tau-1}^{\tau,i}) \right] \\  &\qquad \cdot \sum_{j=1}^N \overline{w}_{t-1}^j \delta_{x_{1:t-1}^{t-1,j}}( x_{1:t-1}^{t}) r_{\tau}(x_{t}^{t}|x_{t-1}^{t}). \notag\\[10pt]
R_{t}(x_{1:1}^{1,1:N},...,x_{1:t-1}^{t-1,1:N},x_{1:t}^t)&=R_{t-1}^N(x_{1:1}^{1,1:N},...,x_{1:t-1}^{t-1,1:N})\cdot \frac{p(x_{1:t}^t,y_{1:t})/p(x_{1:t-1}^t,y_{1:t})}{r_{t}(x_{t}^{t}|x_{t-1}^{t})}\notag\\
&=\prod_{\tau=1}^{t-1}\left[\frac{1}{N}\sum_{i=1}^N w_\tau^i\right] \frac{f(x_{t}^{t}|x_{t-1}^{t})g(y_{t}|x_{t}^{t})}{r_{t}(x_{t}^{t}|x_{t-1}^{t})}.\notag \\[10pt]
a_{t}(\cdot|x_{1:1}^{1,1:N},...,x_{1:t-1}^{t-1,1:N},x_{1:t}^t) & = \delta_{x_{1:t}^{t}}(\cdot).\notag
\end{align*}
By Lemma 3, $(Q_t, R_t, a_t)$ define an estimator-coupling pair for $p(x_{1:t}, y_{1:t})$.

It is now straightforward to verify that applying Lemma 2 give the triple $(Q_t^N, R_t^N, a_t^N)$ defined in Equations~\eqref{eq:Q_SMC_proof}--\eqref{eq:a_SMC_proof}, and therefore $(Q_t^N, R_t^N, a_t^N)$ define an estimator-coupling pair for $p(x_{1:t}, y_{1:t})$, and the result is proved.

\subsection{Proof of Theorem 2}
We again repeat the relevant definitions. The weights used in VMPF are
$$
v_1^i = \frac{f(x_1^i) g(y_1|x_1^i)}{r_1(x_1^i)},
\quad v_t^i = \frac{\sum_{j=1}^Nv_{t-1}^jf(x_t^i|x_{t-1}^j)g(y_t|x_t^i)}{\sum_{j=1}^N v_{t-1}^jr_t(x_t^i | x_{t-1}^j)} \text{ for $t > 1$}.
$$
The normalized weights are $\overline{v}_t^i=v_t^i/\left(\sum_{j=1}^Nv_t^j\right)$.

We wish to show that, for all $t$,
\begin{align}
Q_{t}^N( x_1^{1:N}, \ldots , x_{t}^{1:N}) &= \prod_{i=1}^N \left[r_1(x_1^i) \prod_{\tau=2}^{t} \sum_{j=1}^N \overline{v}_{\tau-1}^j r_{\tau}(x_{\tau}^i|x_{\tau-1}^j) \right], \label{eq:Q_MPF_proof} \\
R_t^N(x_1^{1:N}, \ldots , x_{t}^{1:N}) &= \prod_{\tau=1}^{t}\frac{1}{N} \sum_{i=1}^N v_{\tau}^i, \label{eq:R_MPF_proof} \\
a_t^N(\cdot|x_1^{1:N}, \ldots , x_{t}^{1:N}) &= \sum_{i=1}^N \overline{v}_t^i \delta_{x_t^i}(\cdot) \label{eq:a_MPF_proof}
\end{align}
define an estimator-coupling pair for $p(x_t, y_{1:t})$.

We can check that these match the MPF algorithm at step $t$: that is, $Q_t^N$ is sampling distribution, $R_t^N$ is the likelihood estimator, and $a_t^N$ is approximation to $p(x_t|y_{1:t})$. So, after proving this, the conclusion of the theorem follows immediately.

We will show inductively that $(Q_t^N, R_t^N, a_t^N)$ are obtained by applying appropriate operations on estimator-coupling pairs to. In particular, the procedure is 
\begin{align*}
(Q_t^N, R_t^N, a_t^N) &= \textsc{Replicate}\left(Q_t, R_t, a_t; x_t, N\right)
\end{align*}
for all $t$, where
\begin{align*}
	Q_1(x_{1})&=r_1(x_{1}), \\
	R_1(x_1)&=\frac{p(x_{1},y_1)}{r_1(x_{1})}=\frac{f(x_{1})g(y_1|x_{1})}{r_1(x_{1})}, \\
	a_1(\cdot|x_1)&=\delta_{x_{1}}(\cdot),
\end{align*}
and, for $t > 1$, 
\begin{align*}
(Q'_t, R'_t, a'_t) &= \textsc{ChangeTarget}\Big(Q_{t-1}^N, R_{t-1}^N, a_{t-1}^N; p(x_{t-1}, y_{1:t-1}), p(x_{t-1},x_t, y_{1:t}), r_t(x_t | x_{t-1}) \Big), \\
(Q_t, R_t, a_t) &= \textsc{Marginalize}\Big(Q'_{t}, R'_{t}, a'_{t}; \hat x_{t-1} \Big).
\end{align*}

The proof is again a mechanical application of these operations.

The base case is identical to the base case in the proof of Theorem 1, and yields that $(Q_1^N, R_1^N, a_1^N)$ have the form in Equations~\eqref{eq:Q_MPF_proof}--\eqref{eq:a_MPF_proof} and define an estimator-coupling pair for $p(x_1, y_1)$. 

For the induction step $(t > 1)$, we first apply Lemma 3 as in the proof of Theorem 1, except using the \textsc{ChangeTarget} operation instead of \textsc{ExtendTarget}, to get
\begin{align*}
  Q_t'(x_1^{1:N}, \ldots , x_{t-1}^{1:N}, \hat{x}_{t-1}, x_t)
  &= \prod_{i=1}^N \left[ r_1(x_1^i) \prod_{\tau=2}^{t-1} \sum_{j=1}^N \overline{v}_{\tau-1}^j r_{\tau}(x_{\tau}^i|x_{\tau-1}^j) \right] \\
  &\qquad \cdot \sum_{j=1}^N \overline{v}_{t-1}^j \delta_{x_{t-1}^j} (\hat{x}_{t-1}) r_{t}(x_{t}|\hat{x}_{t-1}), \\
  R_t'(x_1^{1:N}, \ldots, x_{t-1}^{1:N}, \hat{x}_{t-1}, x_t)
  &= \prod_{\tau=1}^{t-1}\left[\frac{1}{N}\sum_{i=1}^Nv_{\tau}^i\right] \frac{f(x_t|\hat{x}_{t-1})g(y_t|x_t)}{r_t(x_t|\hat{x}_{t-1})},\\
  a_t'(\cdot|x_1^{1:N},...,x_{t-1}^{1:N},\hat{x}_{t-1},x_t)&=\delta_{x_t}(\cdot).
\end{align*}
which define an estimator-coupling pair for $p(x_t, y_{1:t})$.

Next, we apply the \textsc{Marginalize} operation using Lemma 1. Marginalizing $\hat{x}_{t-1}$ from $Q_t'$ gives
\begin{align*}
Q_t(x_1^{1:N}&,\ldots,x_{t-1}^{1:N},x_t) \\
	=&\int Q_t'(x_1^{1:N},\ldots,x_{t-1}^{1:N},\hat{x}_{t-1},x_t)d\hat{x}_{t-1} \\
	=&\int \prod_{i=1}^N \left[ r_1(x_1^i) \prod_{\tau=2}^{t-1} \sum_{j=1}^N \overline{v}_{\tau-1}^j r_{\tau}(x_{\tau}^i|x_{\tau-1}^j) \right] \sum_{j=1}^N \overline{v}_{t-1}^j \delta_{x_{t-1}^j}(\hat{x}_{t-1}) r_{t}(x_{t}|\hat{x}_{t-1}) d\hat{x}_{t-1} \\
        =& \prod_{i=1}^N \left[ r_1(x_1^i) \prod_{\tau=2}^{t-1} \sum_{j=1}^N \overline{v}_{\tau-1}^j r_{\tau}(x_{\tau}^i|x_{\tau-1}^j) \right] \sum_{j=1}^N \int \overline{v}_{t-1}^j \delta_{x_{t-1}^j}(\hat{x}_{t-1}) r_{t}(x_{t}|\hat{x}_{t-1}) d\hat{x}_{t-1}\\
        =& \prod_{i=1}^N \left[ r_1(x_1^i) \prod_{\tau=2}^{t-1} \sum_{j=1}^N \overline{v}_{\tau-1}^j r_{\tau}(x_{\tau}^i|x_{\tau-1}^j) \right] \sum_{j=1}^N \overline{v}_{t-1}^j r_{t}(x_{t}|x_{t-1}^j).
\end{align*}

The conditional distribution is
\begin{align*}
Q'_t( \hat x_{t-1} \,|\, x_1^{1:N} ,\ldots,x_{t-1}^{1:N},x_t) 
&= \frac{Q'_t(x_1^{1:N},\ldots,x_{t-1}^{1:N}, \hat x_{t-1}, x_t)}{Q'_t(x_1^{1:N},\ldots,x_{t-1}^{1:N},x_t)} \\
&= \frac
{\sum_{j=1}^N \overline{v}_{t-1}^j \delta_{x_{t-1}^j}( \hat{x}_{t-1}) r_{t}(x_{t}|\hat{x}_{t-1})}
{\sum_{j=1}^N \overline{v}_{t-1}^j r_{t}(x_{t}|x_{t-1}^j)}.
\end{align*}

The new estimator $R_t$ is the conditional expectation
\begin{align*}
  R_t(x_1^{1:N}&, \ldots , x_{t-1}^{1:N},x_t) \\
  &= \EE_{Q_t'(\hat{x}_{t-1}|x_1^{1:N}, \ldots , x_{t-1}^{1:N}, x_t)} R_t'(x_1^{1:N}, \ldots, x_{t-1}^{1:N}, \hat{x}_{t-1}, x_t) \\
  &= \int \frac
    {\sum_{j=1}^N \overline{v}_{t-1}^j \delta_{x_{t-1}^j}( \hat{x}_{t-1}) r_{t}(x_{t}|\hat{x}_{t-1})}
    {\sum_{j=1}^N \overline{v}_{t-1}^j r_{t}(x_{t}|x_{t-1}^j)}
    \prod_{\tau=1}^{t-1}\left[\frac{1}{N}\sum_{i=1}^Nv_{\tau}^i\right]
    \frac{f(x_t|\hat{x}_{t-1})g(y_t|x_t)}{r_t(x_t|\hat{x}_{t-1})}
    d\hat x_{t-1} \\
   &= \frac{1} { \sum_{j=1}^N \overline{v}_{t-1}^j r_{t}(x_{t}^i|x_{t-1}^j)} \prod_{\tau=1}^{t-1} \left[ \frac{1}{N} \sum_{i=1}^N v_{\tau}^i \right] \int \sum_{j=1}^N \overline{v}_{t-1}^j \delta_{x_{t-1}^j}(\hat{x}_{t-1}) f(x_t|\hat{x}_{t-1}) g(y_t|x_t) d\hat{x}_{t-1}\\	&= \prod_{\tau=1}^{t-1} \left[ \frac{1}{N} \sum_{i=1}^N v_{\tau}^i \right] \frac{ \sum_{j=1}^N v_{t-1}^j f(x_t|x_{t-1}^j) g(y_t|x_t)}{\sum_{j=1}^N v_{t-1}^j r_{t}(x_{t}^i|x_{t-1}^j)}.
\end{align*}

The new coupling $a_t$ is
\begin{align*}
a_t(x_1^{1:N}&, \ldots, x_{t-1}^{1:N}, x_t) \\
&= \frac{1}{ R_t(x_1^{1:N}, \ldots, x_{t-1}^{1:N}, x_t)} \\
&\qquad \cdot \EE_{Q_t'(\hat{x}_{t-1}| x_1^{1:N}, \ldots, x_{t-1}^{1:N}, x_t)} \left[ R_t'(x_1^{1:N}, \ldots , x_{t-1}^{1:N}, \hat{x}_{t-1}, x_t) a_t'(\cdot| x_1^{1:N}, \ldots , x_{t-1}^{1:N}, \hat{x}_{t-1}, x_t)\right] \\
&= \frac{1}{R_t(x_1^{1:N}, \ldots,x_{t-1}^{1:N},x_t)} \EE_{Q_t'(\hat{x}_{t-1}|x_1^{1:N},...,x_{t-1}^{1:N},x_t)} \left[R_t'(x_1^{1:N},...,x_{t-1}^{1:N},\hat{x}_{t-1},x_t)\right] \delta_{x_t}(\cdot)\\[3pt]
&= \delta_{x_t}(\cdot).
\end{align*}

It is now straightforward to verify that applying Lemma 2 give the triple $(Q_t^N, R_t^N, a_t^N)$ defined in Equations~\eqref{eq:Q_MPF_proof}--\eqref{eq:a_MPF_proof}, and therefore $(Q_t^N, R_t^N, a_t^N)$ define an estimator-coupling pair for $p(x_{t}, y_{1:t})$, and the result is proved.

\section{Deriving MPF from IPF}
\label{sec:MPFfromIPF}
We directly give the detail of IPF in Algorithm \ref{algorithm3}.
\begin{figure}[t]
  \centering
  \begin{minipage}{.48\linewidth}
    \begin{algorithm}[H]
		\caption{Independent Particle Filters} 
		\label{algorithm3}
		\begin{algorithmic}[1]
			\Require $p(x_{1:T},y_{1:T})$, $y_{1:T}$, $\{r_t(x_t)\}$,  $N$
                        \State Sample $x^{i}_1 \sim r_1(x_1)$ for all $i$
                        \State Set $u_1^i = \frac{f(x_1^{i})g(y_1|x_1^{i})}{r_1(x_1^{i})}$ for all $i$

			\For {$t=2,\ldots,T$}
			      \State Generate $L$ permutations $k_{t}^{(1:L,1:N)}$
                  \For {$i=1,\ldots,N$}
                  		\State Sample $x_t^i\sim r_t(x_t)$

                        \State Set $u_t^i = \frac{\sum_{l=1}^Lu_{t-1}^{k_{l,i}}f(x_t^i|x_{t-1}^{k_{l,i}})g(y_t|x_t^i)}{L\cdot r_t(x_t^i)}$
                  \EndFor     	
			\EndFor
		\end{algorithmic}
	\end{algorithm}
  \end{minipage}
  \hfill
    \begin{minipage}{.48\linewidth}
    \begin{algorithm}[H]
		\caption{Tensor Monte Carlo for SSM} 
		\label{algorithm4}
		\begin{algorithmic}[1]
			\Require $p(x_{1:T},y_{1:T})$, $y_{1:T}$, $\{r_t(x_t)\}$,  $N$
			\State Sample $x^{i}_1 \sim r_1(x_1)$ for all $i$
            \State Set $z_1^i = \frac{f(x_1^{i})g(y_1|x_1^{i})}{r_1(x_1^{i})}$ for all $i$
			\For {$t=2,\ldots,T$}
                  \For {$i=1,\ldots,N$}
                      \State Sample $x_t^i\sim r_t(x_t)$
                      \State Set $z_t^i=\frac{\sum_{j=1}^Nz_{t-1}^jf(x_t^i|x_{t-1}^j)g(y_t|x_t^i)}{N\cdot r_t(x_t^i)}$
                  \EndFor
            \EndFor
		\end{algorithmic}
	\end{algorithm}
  \end{minipage}
\end{figure}
In line 4 of Algorithm \ref{algorithm3}, $L$ distinct permutations ($k_{t}^{l_1,i}\neq k_t^{l_2,i}$ for any $l_1,l_2,i$ satisfying $l_1\neq l_2$) of $1,\ldots,N$ should be generated. Proposition 1 of IPF \citep{lin2005independent} implies that for any test function $h$,
\begin{align}
\label{eq:IPFunbiasedness}
	\mathbb{E}\left[\frac{1}{N}\sum_{i=1}^Nu_t^ih(x_t^i)\right]=p(y_{1:t})\cdot \mathbb{E}_{p(x_t|y_{1:t})}\left[h(x_t)\right].
\end{align}
If $h \equiv 1$, we have $\mathbb{E}\left[\frac{1}{N}\sum_{i=1}^Nu_t^i\right]=p(y_{1:t})$. 

The idea of TMC is to draw many copies of each individual variable from independent proposal distributions, and then average over all (exponentially many) combinations of joint samples to get an unbiased estimator. We assume for each latent variable $x_t$, we have $N$ samples $x_t^i\sim r_t(x_t)$, $i=1,\ldots, N$. So
\begin{align}
\label{eq:TMCestimator}
	\hat{p}_{TMC}&=\frac{1}{N^T}\sum_{i_1,i_2,...,i_T}\frac{p(x_1^{i_1},\ldots,x_T^{i_T},y_{1:T})}{r_1(x_1^{i_1})\ldots r_T(x_T^{i_T})}\notag\\
	&=\frac{1}{N^T}\sum_{i_1,i_2,...,i_T}\frac{f(x_1^{i_1})g(y_1|x_1^{i_1})}{r_1(x_1^{i_1})}\cdot \frac{f(x_2^{i_2}|x_1^{i_1})g(y_2|x_2^{i_2})}{r_2(x_2^{i_2})}\cdot \ldots\cdot \frac{f(x_T^{i_T}|x_{T-1}^{i_{T-1}})g(y_T|x_T^{i_T})}{r_T(x_T^{i_T})}
\end{align}
will be an unbiased estimator for $p(y_{1:T})$. Computing the summation in Equation (\ref{eq:TMCestimator}) can be accelerated by summing over $i_1$, $i_2$, ..., $i_T$ in order. If we define $z_1^i=\frac{f_1(x_1^{i})g(y_1|x_1^{i})}{r_1(x_1^{i})}$, then summing over $i_1$ gives
\begin{align}
	\label{eq:TMCderivation}
	\hat{p}_{TMC}&=\frac{1}{N^T}\sum_{i_1,i_2,...,i_T}z_1^{i_1}\cdot \frac{f(x_2^{i_2}|x_1^{i_1})g(y_2|x_2^{i_2})}{r_2(x_2^{i_2})}\cdot \ldots\cdot \frac{f(x_T^{i_T}|x_{T-1}^{i_{T-1}})g(y_T|x_T^{i_T})}{r_T(x_T^{i_T})}\notag\\
	&=\frac{1}{N^{T-1}}\sum_{i_2,...,i_T} \boxed{\sum_{i_1=1}^{N}\frac{z_1^{i_1}f(x_2^{i_2}|x_1^{i_1})g(y_2|x_2^{i_2})}{N\cdot r_2(x_2^{i_2})}}\cdot \ldots\cdot \frac{f(x_T^{i_T}|x_{T-1}^{i_{T-1}})g(y_T|x_T^{i_T})}{r_T(x_T^{i_T})}.
\end{align}
We can further define the boxed variable in Equation (\ref{eq:TMCderivation}) by $z_2^{i_2}$ and continue the summation. This gives a filtering style framework as in Algorithm \ref{algorithm4}. In the end, $\frac{1}{N}\sum_{i=1}^N z_t^i$ will also be an unbiased estimator for $p(y_{1:t})$. 
\subsection{Deriving TMC for SSM from IPF}
Now we show that TMC is IPF with complete matching ($L=N$). Under this circumstance, $k_t^{1,i},\ldots,k_t^{L,i}$ will also be a permutation of $1,\ldots, N$. So Line 7 of Algorithm \ref{algorithm3} will become
\begin{align}
	u_t^i = \frac{\sum_{j=1}^Nu_{t-1}^{j}f(x_t^i|x_{t-1}^{j})g(y_t|x_t^i)}{N\cdot r_t(x_t^i)}\notag,
\end{align}
which exactly matches Line 6 of Algorithm \ref{algorithm4}.
\subsection{Deriving MPF from TMC for SSM}
The proposal distributions of IPF and TMC can be extended to condition on all past particles, i.e. $r_t=r_t(x_t|x_{1:t-1}^{1:N})$ (See 7.2 of \citet{Aitchison19} about non-factorized TMC), which includes the case of MPF. However, no formal theory supports any specific form of proposal distribution, such as the mixture distribution used by MPF. Additionally, full details of how to  construct the non-factorised approximate posterior and ensure differentiability are not given. Finally, by using couplings to derive MPF and VMPF, we obtain the interpretation as auxiliary variable VI as described in the main text, which is not obvious from the unbiasedness result of Equation (\ref{eq:IPFunbiasedness}) alone.



This paper fills these gaps by deriving MPF from SMC directly and revealing the KL decomposition for VMPF. Now we verify that MPF is TMC with a specific mixture distribution. If we define $v_t^i=z_t^i/\left(\prod_{\tau=1}^{t-1}\frac{1}{N}\sum_{j=1}^Nv_{\tau}^j\right)$ and replace the proposal distributions for $t>1$ by $\sum_{j=1}^N\overline{v}_{t-1}^jr_t(x_t^i|x_{t-1}^j)$, we will have $v_1^i=z_1^i$ for all $i$ and
\begin{align}
	&z_t^i=\frac{\sum_{j=1}^Nz_{t-1}^jf(x_t^i|x_{t-1}^j)g(y_t|x_t^i)}{N\cdot \sum_{j=1}^N\overline{v}_{t-1}^jr_t(x_t^i|x_{t-1}^j)}\notag\\
	\Longleftrightarrow & \left(\prod_{\tau=1}^{t-1}\frac{1}{N}\sum_{j=1}^Nv_{\tau}^j\right)v_t^i=\left(\prod_{\tau=1}^{t-2}\frac{1}{N}\sum_{j=1}^Nv_{\tau}^j\right)\frac{\sum_{j=1}^Nv_{t-1}^jf(x_t^i|x_{t-1}^j)g(y_t|x_t^i)}{N\cdot \sum_{j=1}^N\overline{v}_{t-1}^jr_t(x_t^i|x_{t-1}^j)}\notag\\
	\Longleftrightarrow & \left(\prod_{\tau=1}^{t-1}\frac{1}{N}\sum_{j=1}^Nv_{\tau}^j\right)v_t^i=\left(\prod_{\tau=1}^{t-1}\frac{1}{N}\sum_{j=1}^Nv_{\tau}^j\right)\frac{\sum_{j=1}^Nv_{t-1}^jf(x_t^i|x_{t-1}^j)g(y_t|x_t^i)}{\sum_{j=1}^N v_{t-1}^jr_t(x_t^i|x_{t-1}^j)}\notag\\
	\Longleftrightarrow & v_t^i=\frac{\sum_{j=1}^Nv_{t-1}^jf(x_t^i|x_{t-1}^j)g(y_t|x_t^i)}{\sum_{j=1}^N v_{t-1}^jr_t(x_t^i|x_{t-1}^j)}.
\end{align}
Therefore, the weight computation of MPF is equivalent to Line 6 of Algorithm \ref{algorithm4} if the specific mixture distribution is chosen. And we directly have the unbiased estimator
\begin{align}
	\frac{1}{N}\sum_{i=1}^N z_t^i=\prod_{\tau=1}^{t}\frac{1}{N}\sum_{i=1}^Nv_{\tau}^i 
\end{align}
to be in the same form as MPF. 
\section{Limitation on VRNN}
A VRNN is a sequential latent variable model 
\begin{align}
  p(x_{1:T},y_{1:T})=\prod_{t=1}^Tf(x_t|h_t)g(y_t|x_t,h_t)\notag
\end{align}
where $h_1$ is constant and $h_t=h(x_{t-1},y_{t-1},h_{t-1})$ for $t>1$. The variational posterior distribution is factorized as 
\begin{align}
  q(x_{1:T}|y_{1:T})=\prod_{t=1}^Tq_t(x_t|h_t,y_t).\notag
\end{align}
We interpret a VRNN as a SSM that has both deterministic latent variables $h_t$ and stochastic latent variables $x_t$. The full model is 
\begin{align}
	p(x_{1:T}, h_{2:T},y_{1:T})=\prod_{t=1}^Tf(x_t|h_t)g(y_t|x_t,h_t)\prod_{t=2}^T\delta_{h(x_{t-1},y_{t-1},h_{t-1})}(h_t).\notag
\end{align}
To be clearer, we define $X_t=(x_t,h_t)$ to be the full latent variable, so the model becomes $p(X_{1:T},y_{1:T})$. VSMC can be used to learn the parameters of a VRNN. However, VMPF is not applicable to learning VRNN parameters. If we set the proposal distribution to be 
\begin{align}
	r_t(X_t|X_{t-1},y_{1:t})=\delta_{h(x_{t-1},y_{t-1},h_{t-1})}(h_t)q_t(x_t|h_t,y_t),\notag
\end{align}
the weighting function for VMPF for $t>1$ would become
\begin{align}
	v_t^i=\frac{\sum_{j=1}^Nv_{t-1}^j\delta_{h(x_{t-1}^j,y_{t-1},h_{t-1}^j)}(h_t^i)f(x_t^i|h_t^i)g(y_t|x_t^i,h_t^i)}{\sum_{j=1}^Nv_{t-1}^j\delta_{h(x_{t-1}^j,y_{t-1},h_{t-1}^j)}(h_t^i)q_t(x_t^i|h_t,y_t)}.\notag
\end{align}
We know that, in general, for all $j$, the values $h(x_{t-1}^j,y_{t-1},h_{t-1}^j)$ will be different from each other. Due to the presence of Dirac function, the summation in the numerator as well as the denominator will only have one non-zero term, so VMPF collapses to VSMC. This failure can be understood in terms of marginalization: if we are able to access $h_t^i$ with MPF at $t$, we can reconstruct the corresponding $(x_{t-1}^j,h_{t-1}^j)$ that generates $(x_{t}^i,h_{t}^i)$, but $(x_{t-1}^j,h_{t-1}^j)$ should be marginalized.
\section{Experiment Details}
We run all experiments on CPU. For models with multiple data, the batch size is 1. The experiment details can be found below.
\subsection{Details of experiments on stochastic volatility models}
Suppose the raw data of the exchange rate is $d_{0:T}=(d_0,d_1,...,d_T)$, we get $y_{1:T}$ by
\begin{align}
	y_t=\log d_t-\log d_{t-1}\notag
\end{align}
for $1\le t\le T$. For both diagonal $B$ and triangular $B$ case, we restrict elements of $\Phi$ to be in $[0,1]$. For all algorithms, we train with the same learning rate scheduler (See Table $\ref{TrainingScheduleSVModel}$), except for VMPF-UG for diagonal $B$ with $N=16$, where we train with 0.3x of the listed learning rate. Further reducing the learning rate has little effect on the results.
 To stabilize the training of VMPF-UG, we use gradient clipping with threshold $100$. With that, VMPF-UG for diagonal $B$ with $N=8,16$ are not stable in early iterations, and we only keep stable runs. 
\begin{table}[t]
\centering
\caption{Training schedule on stochastic volatility model.}
\label{TrainingScheduleSVModel}
\begin{tabular}{ccccc}
&\multicolumn{2}{c}{Diagonal $B$}&\multicolumn{2}{c}{Triangular $B$}\\
Phase&Learning Rate&Epochs&Learning Rate&Epochs\\
\toprule
1&0.01&50000&0.003&100000\\
2&0.001&50000&0.0003&100000\\
3&0.0001&50000&0.00003&100000\\
4&0.00001&50000&0.000003&100000\\
 \bottomrule
	\end{tabular}
\end{table}
\subsection{Details of experiments on deep Markov models}
The four polyphonic music datasets are sequences of 88-dimensional binary vectors. Recall that the DMM is
\begin{align}
	x_t&=\mu_{\theta}(x_{t-1}) + \mathrm{diag}(\exp(\sigma_{\theta}(x_{t-1})/2))v_t,\notag\\
	y_t&\sim \mathrm{Bernoulli}(\mathrm{sigmoid}(\eta_{\theta}(x_t))),\notag
\end{align}
where $v_t\sim \mathcal{N}(0,I)$, $x_0=0$, and $\mu_\theta$, $\sigma_\theta$, $\eta_\theta$ are neural networks. The architectures are
\begin{align}
	\mu_{\theta},\sigma_\theta &=\mathrm{Linear}(d_h\to 176)\circ \mathrm{LeakyReLU}\circ \mathrm{Linear}(88\to d_h),\notag\\
	\eta_\theta &=\mathrm{Linear}(d_h\to 88)\circ \mathrm{LeakyReLU}\circ \mathrm{Linear}(88\to d_h)\notag
\end{align}
where $d_h$ varies on different datasets. And we use the proposal distribution
\begin{align}
  r(x_t|x_{t-1},y_t;\phi)\propto \mathcal{N}\big(x_t;\mu^x_{\phi}(x_{t-1}),\diag(\exp(\sigma^x_{\phi}(x_{t-1})))\big)
  \cdot \mathcal{N}\big(x_t;\mu^y_{\phi}(y_t),\diag(\exp(\sigma^y_{\phi}(y_t)))\big)\notag
\end{align}
where $\mu^x_\phi$, $\sigma^x_\phi$, $\mu^y_\phi$ and $\sigma^y_\phi$ are neural networks. The architectures are
\begin{align}
	\mu^x_\phi, \sigma^x_\phi &=\mathrm{Linear}(d_h\to 176)\circ \mathrm{LeakyReLU}\circ \mathrm{Linear}(88\to d_h),\notag\\
	\mu^y_\phi, \sigma^y_\phi &= \mathrm{Linear}(d_h\to 176)\circ \mathrm{LeakyReLU}\circ \mathrm{Linear}(88\to d_h).\notag
\end{align}
For training, we use the default train, validation, and test set split. For all datasets, the learning rate, training epochs and hidden units $h_t$ can be found in Table \ref{TrainingDetailsDMM}. The comma separates different phases of training: we train with the larger learning rates for some epochs and reduce the learning rate for some additional epochs. We report the performance on test set with the best parameters during training determined by the validation set. 
\begin{table}[t]
\centering
\caption{Training details of DMMs.}
\label{TrainingDetailsDMM}
\begin{tabular}{ccccc}
&Nottingham&JSB&MuseData&Piano-midi.de\\
\toprule
Learning Rate &0.001,0.0001&0.001,0.0001&0.001,0.0001&0.001,0.0001\\
Epochs &500,100&1000,200&350,100&600,150\\
Hidden Units $d_h$&128&64&128&128\\
 \bottomrule
	\end{tabular}
\end{table}

\end{document}